\documentclass[10pt,journal,compsoc]{IEEEtran}
\usepackage{subfigure}
\usepackage{graphicx}
\usepackage{caption} 
\usepackage{amsfonts,amssymb, amsthm, amsmath}
\usepackage{rotating}
\usepackage{multirow, multicol}
\usepackage{makecell}
\usepackage{diagbox}
\usepackage{soul}
\usepackage[linesnumbered,ruled, vlined]{algorithm2e}
\usepackage{algpseudocode}
\usepackage{arydshln} 
\usepackage[usenames,dvipsnames]{xcolor} 
\usepackage[pagebackref,breaklinks,colorlinks]{hyperref}
\usepackage[capitalize]{cleveref}
\usepackage{caption}

\crefname{section}{Sec.}{Secs.}
\Crefname{section}{Section}{Sections}
\Crefname{table}{Table}{Tables}
\crefname{table}{Tab.}{Tabs.}

\newcommand{\Tref}[1]{Table~\ref{#1}}
\newcommand{\Eref}[1]{Eq.~(\ref{#1})}
\newcommand{\Fref}[1]{Fig.~\ref{#1}}
\newcommand{\Sref}[1]{Section~\ref{#1}}

\newcommand{\mycolor}[0]{\color{black}}
\newcommand{\mycaption}[1]{\captionsetup{font={color=black}}\caption{#1}}
\newcommand{\qiankun}[1]{\color{black}#1\color{black}}
\newcommand{\revise}[1]{\color{black}#1\color{black}}

\ifCLASSOPTIONcompsoc
\usepackage[nocompress]{cite}
\else
\usepackage{cite}
\fi
\ifCLASSINFOpdf
\else
\fi
\begin{document}
		%
		\title{Transformer based Pluralistic Image Completion \\ with Reduced Information Loss}
		%
		%
		%
		%
		
		\author{Qiankun Liu, Yuqi Jiang, Zhentao Tan, Dongdong Chen$^\dag$,  Ying Fu$^{\dag}$,~\IEEEmembership{Senior Member, IEEE}, \\ Qi Chu$^\dag$, Gang Hua, ~\IEEEmembership{Fellow,~IEEE}, Nenghai Yu
			\IEEEcompsocitemizethanks{
				\IEEEcompsocthanksitem Qiankun Liu, Yuqi Jiang and Ying Fu are with the School of Computer Science and Technology, Beijing Institute of Technology (\{liuqk3, yqjiang, fuying\}@bit.edu.cn).
				\IEEEcompsocthanksitem Zhentao Tan, Qi Chu and Nenghai Yu are with the School of Information Science and Technology, University of Science and Technology of China ( tzt@mail.ustc.edu.cn; \{qchu, ynh\}@ustc.edu.cn).
				\IEEEcompsocthanksitem Dongdong Chen is with Microsoft Cloud AI (cddlyf@gmail.com).
				\IEEEcompsocthanksitem Gang Hua is with Wormpex AI Research LLC (ganghua@gmail.com).
				\IEEEcompsocthanksitem Dongdong Chen, Ying Fu and Qi Chu are co-corresponding authors.
			}
		}
		
		%
		%

	\markboth{IEEE TRANSACTIONS ON PATTERN ANALYSIS AND MACHINE INTELLIGENCE}%
	{Liu \MakeLowercase{\textit{et al.}}: Transformer based Pluralistic Image Completion with Reduced Information Loss}
	%



	\IEEEtitleabstractindextext{%
		\begin{abstract}
			Transformer based methods have achieved great success in image inpainting recently. However, we find that these solutions regard each pixel as a token, thus suffering from an information loss issue from two aspects: 1) They downsample the input image into much lower resolutions for efficiency consideration. 2) They quantize $256^3$ RGB values to a small number (such as 512) of quantized color values. The indices of quantized pixels are used as tokens for the inputs and prediction targets of the transformer. To mitigate these issues, we propose a new transformer based framework called ``\textbf{PUT}''. Specifically, to avoid input downsampling while maintaining computation efficiency, we design a patch-based auto-encoder \textbf{P}-VQVAE. The encoder converts the masked image into non-overlapped patch tokens and the decoder recovers the masked regions from the inpainted tokens while keeping the unmasked regions unchanged. To eliminate the information loss caused by input quantization, an \textbf{U}n-quantized \textbf{T}ransformer is applied. It directly takes features from the P-VQVAE encoder as input without any quantization and only regards the quantized tokens as prediction targets. \qiankun{Furthermore, to make the inpainting process more controllable, we introduce semantic and structural conditions as extra guidance.} Extensive experiments show that our method greatly outperforms existing transformer based methods on image fidelity and achieves much higher diversity and better fidelity than state-of-the-art pluralistic inpainting methods on complex large-scale datasets (\textit{e.g.}, ImageNet). Codes are available at \href{https://github.com/liuqk3/PUT}{https://github.com/liuqk3/PUT}.
		\end{abstract}
		
		}

	\maketitle

	\IEEEdisplaynontitleabstractindextext

	%
	\IEEEpeerreviewmaketitle

	\IEEEraisesectionheading{
		\section{Introduction}
		\label{sec:intro}
	}
	\IEEEPARstart{I}{mage} inpainting, which focuses on filling meaningful and plausible contents in missing regions for damaged images, is a hot topic in the computer vision field and widely used in various real applications~\cite{qiu2020semanticadv, zhan2020self, barnes2009patchmatch, tan2021efficient, matsushita2006full}. Traditional methods~\cite{bertalmio2003simultaneous, barnes2009patchmatch, criminisi2004region} based on texture matching can handle simple cases very well but struggle for complex natural images. In the last several years, benefiting from the development of CNNs, tremendous success~\cite{liu2018image,nazeri2019edgeconnect} has been achieved by learning on large-scale datasets. However, due to the inherent properties of CNNs, \textit{i.e.}, local inductive bias and spatial-invariant kernels, such methods still do not perform well in understanding global structure and inpainting large masked/missing regions.

	\begin{figure}[t]
		\centering
		\includegraphics[width=1.0\columnwidth]{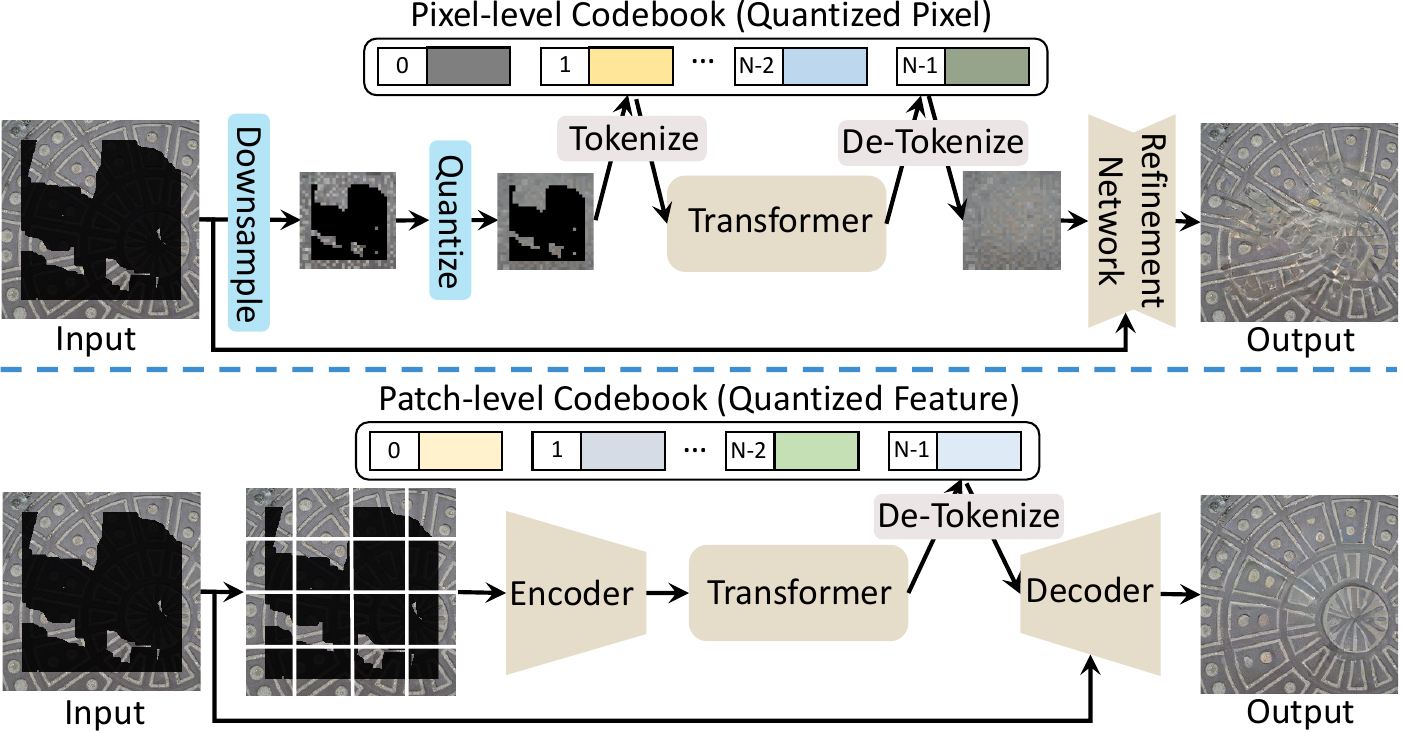} 
		\caption{Top: Existing transformer based autoregressive methods~\cite{wan2021high, yu2021diverse}. The output is produced by ICT~\cite{wan2021high}. Bottom: Our transformer based method. \emph{``Tokenize''} means getting the indices of quantized pixels or latent vectors, and \emph{``De-Tokenize''} is the inverse operation.
		}
		\label{figure: different_type_of_methods}
	\end{figure}

	Recently, transformers have demonstrated their power in various vision tasks~\cite{carion2020end,dong2021cswin, chen2020generative, esser2021taming, ramesh2021zero}, thanks to their capability of modeling long-term relationships. Some recent works~\cite{wan2021high,yu2021diverse} also attempt to apply transformers for image inpainting and have achieved remarkable success in better fidelity and large region inpainting quality. In addition, pluralistic images can be produced when the content of masked regions is predicted and sampled in an autoregressive manner. As shown in the top row of \Fref{figure: different_type_of_methods}, they follow a similar design: 1) Downsample the input image into lower resolutions and quantize the pixels. 2) Use the transformer to recover masked pixels by regarding each quantized pixel as the token. 3) Upsample and refine the low-resolution result by feeding it together with the original input image into an extra CNN network.

	In this paper, we argue that using the above pixel-based token makes existing transformer based autoregressive solutions suffer from the information loss issue from two aspects: 1) \emph{Downsampling}. To avoid the high computation complexity of the transformer, the input image is downsampled into a much lower resolution to reduce the input token number. 2) \emph{Quantization}. To constrain the prediction within a small space, the huge amount ($256^3$, in detail) of RGB pixel values are quantized into much fewer (\textit{e.g.}, 512) quantized pixel values through clustering. The indices of quantized pixels are used as discrete tokens both for the input and prediction target of the transformer. Using the quantized input inevitably further results in information loss.

	\begin{figure*}[t]
		\centering
		\includegraphics[width=2.05\columnwidth]{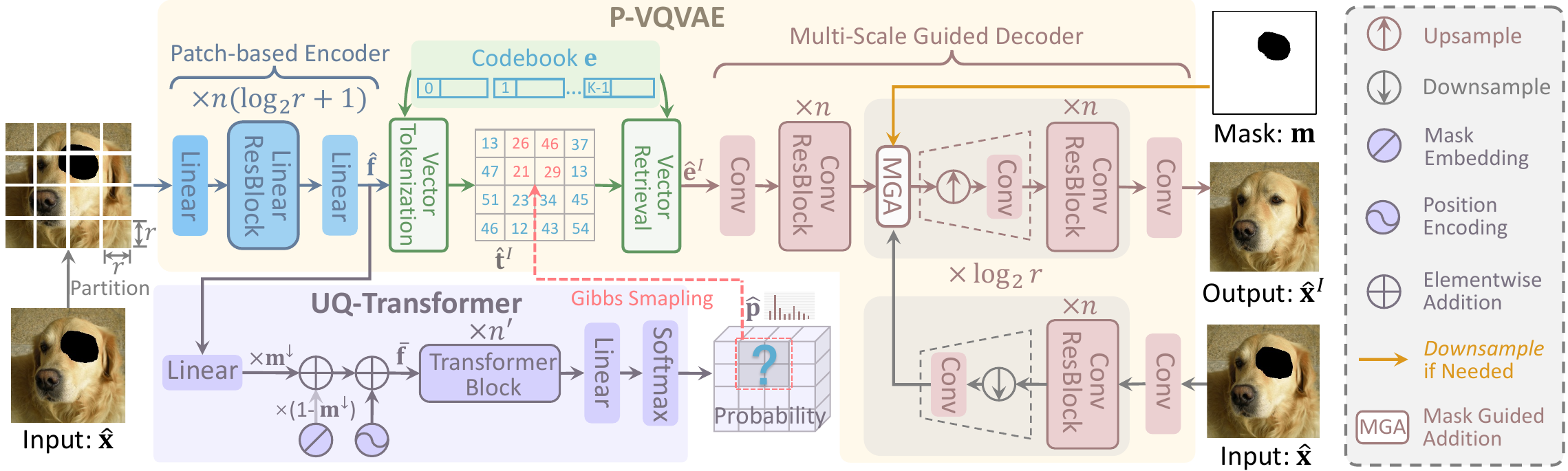} 
		\mycaption{Pipeline of PUT for pluralistic image inpainting without condition. 
			The input image is divided into non-overlapped patches which are processed by the encoder independently. The unmasked regions are reconstructed by the decoder, while the masked regions are predicted by the transformer.}
		\label{figure: transformer_framework}
	\end{figure*}

	To mitigate the aforementioned issues, we propose a new transformer based autoregressive framework \textbf{``PUT"}, which is designed to reduce the information loss in existing inpainting transformers as much as possible. As shown in the bottom row of \Fref{figure: different_type_of_methods}, the original high-resolution input image is directly fed into a \emph{patch-based encoder without any downsampling}, and the transformer directly takes the features from the encoder as \emph{input without any quantization}. 
	
	Specifically, \textbf{PUT} contains two key designs: Patch-based Vector Quantized Variational Auto-Encoder (``\textit{P-VQVAE}”) and Un-Quantized Transformer(``\textit{UQ-Transformer}“). \textit{P-VQVAE} is a dedicatedly designed auto-encoder: 1) Its encoder converts each image patch into a feature vector in a non-overlapped way, where the non-overlapped design is to avoid the disturbance between masked and unmasked regions. 2) As the prediction space of UQ-Transformer, a dual-codebook is built for feature tokenization, where masked and unmasked patches are separately represented by different codebooks. 3) The decoder in P-VQVAE not only recovers the masked image regions from the inpainted tokens but also maintains unmasked regions unchanged. For \textit{UQ-Transformer}, it utilizes the quantized tokens as the prediction targets for masked patches but takes the un-quantized features from the encoder as input. Compared to taking the quantized discrete tokens as input, this design avoids information loss. \qiankun{Furthermore, to make the prediction process controllable, we allow end-users to provide semantic and structural conditions as extra guidance. } 
	Extensive experiments are conducted on FFHQ~\cite{karras2019style}, Places2~\cite{zhou2017places}, and ImageNet~\cite{deng2009imagenet} to demonstrate the superiority of the proposed method. Benefiting from less information loss, PUT achieves much higher fidelity than existing transformer based autoregressive solutions and outperforms state-of-the-art pluralistic inpainting methods by a large margin in terms of diversity.

	\qiankun{This work builds upon our previous CVPR paper. Compared with the conference paper, the main contributions in this paper include:
	\begin{itemize}
		\item We introduce a simple but effective mask embedding for UQ-Transformer to reduce the inpainted artifacts produced by the conference version method. 
		\item We design a multi-token sampling strategy, which enables PUT to take 20× less inference time than the vanilla per-token sampling strategy. 
		\item We propose to integrate the semantic and structural conditions into the generation process, making the final inpainting results more user-controllable.
	\end{itemize}}
  
  \qiankun{In addition, more analysis experiments are conducted, including learning more latent vectors with Gumbel-softmax relaxation strategy, applying PUT to higher resolution images (512×512), controllable image inpainting with extra user guidance, comparing the number of FLOPs, parameters and inference time of different methods, analyzing the impact of the sampling strategy and mask embedding on the quality of inpainted images, and applying pretrained PUT to downstream tasks.}

	\section{Related Work}
	\label{sec:related work}
	
	\subsection{Auto-Encoders}
	Auto-encoders~\cite{hinton1994autoencoders} are a kind of network in unsupervised and semi-supervised learning, which can be divided into three categories: contractive auto-encoder (CAE)~\cite{rifai2011higher}, regularized auto-encoder (RAE)~\cite{min2005non}, and variational auto-encoder (VAE)~\cite{kingma2013auto}. Among them, VAE is the generation model and has become one of the popular frameworks for image synthesis. With the designed two mapping modules between real data and latent space, called encoder and decoder, VAE can be trained in a self-supervised strategy. Diverse images can be generated through the decoder in VAE with latent space sampling or autoregressive models~\cite{van2016pixel,oord2016conditional}. Later, vector quantized variational auto-encoder (VQ-VAE)~\cite{van2017neural,gu2022vector} is proposed for discrete representation learning to circumvent issues of ``posterior collapse'', and further developed by VQ-VAE-2~\cite{razavi2019generating}. Recently, based on the similar quantization mechanism with VQ-VAE, VAGAN~\cite{esser2021taming} and dVAE~\cite{ramesh2021zero} are proposed for conditional image generation through transformers~\cite{vaswani2017attention}. Different from previous methods, the proposed P-VQVAE is dedicated to image inpainting to avoid the disturbance between masked and unmasked regions but with the capability of image reconstruction.

	\subsection{Vision Transformers}
	Thanks to the capability of modeling long-range relationships, transformers have been widely used in different vision tasks, such as image recognition~\cite{dong2021cswin, dong2023peco}, image synthesis~\cite{chen2020generative, esser2021taming, ramesh2021zero}, action recognition~\cite{wang2022bevt}, image inpainting~\cite{wan2021high,yu2021diverse}, and multi-modality learning \cite{yuan2021florence,wang2022omnivl}. Specifically, the autoregressive inference mechanism is naturally suitable for image synthesis related tasks, which can bring diverse results while guaranteeing the quality of the synthesized images~\cite{chen2020generative,esser2021taming,ramesh2021zero,wan2021high}. However, these methods use discrete tokens as the input of transformers. Such practice has a limited effect on image synthesis task but has a non-negligible impact on image inpainting task. In this paper, we propose to replace discrete tokens with continuous feature vectors to avoid information loss.

	Recently, Masked Image Modeling (MIM) methods~\cite{dong2023peco,dong2022bootstrapped,bao2021beit, he2022masked,li2023mage} are proposed for the pretraining of vision transformers with a similar pretext task of image inpainting. However, these methods aim at learning useful representations from incomplete images, leaving the visual quality unpleasant. Different from MIM models, PUT focuses on generating high-quality diverse images with the arbitrary shapes of masks. \qiankun{Nonetheless, the pretrained transformer in PUT learns useful priors for downstream tasks.}

	\subsection{Image Inpainting}
	According to the diversity of inpainted images, there are two different types of definitions for image inpainting task: deterministic image inpainting and pluralistic image inpainting. Most of the traditional methods, whether diffusion-based methods~\cite{bertalmio2000image,efros2001image} or patch-based methods~\cite{barnes2009patchmatch,hays2007scene}, can only generate a single result for each input and may fail while meeting large areas of missing pixels. Later, some CNN based methods~\cite{liu2018image,nazeri2019edgeconnect,suvorov2022resolution,liu2022partial} are proposed to ensure consistency of the semantic content of the inpainted images but still ignore the diversity of results. PICNet~\cite{zheng2019pluralistic} and UCTGAN~\cite{zhao2020uctgan} generate diverse results from the modeled distribution for the masked images. Recently, CoModGAN~\cite{zhao2021comodgan} and MAT~\cite{li2022mat} follow the model structure of StyleGAN2~\cite{karras2020analyzing} to get diverse results with sampled noise and achieve impressive fidelity on monotonous datasets, \textit{e.g.}, FFHQ~\cite{karras2019style} for faces. However, the diversity of these two methods is very limited and they fail to get reasonable results on complex datasets like ImageNet~\cite{deng2009imagenet}. Different from the above-mentioned methods, transformer based solutions~\cite{wan2021high,yu2021diverse} produce diverse results by sampling the tokens of masked patches iteratively in an autoregressive manner. However, their unreasonable design, like the downsampling of input images and quantization of transformer inputs, results in a serious information loss issue. Thus, we propose a novel framework to maximize the input information to achieve better inpainted results. \qiankun{Furthermore, we introduce two types of conditions to make the inpainted results controllable for users.}

	\section{Method}
	\label{sec: method}	
	Our PUT mainly consists of a Patch-based Vector Quantized Variational Auto-Encoder (P-VQVAE) and an Un-Quantized transformer (UQ-Transformer). The overview of our method is shown in \Fref{figure: transformer_framework}. Let ${\mathbf x} \in \mathbb{R}^{H \times W \times 3}$ be an image and ${\mathbf m} \in \{0,1\}^{H \times W \times 1}$ be the mask denoting whether a pixel needs to be inpainted (with value 0) or not (with value 1). $H$ and $W$ are the spatial resolution. The image $\mathbf{\hat{x}}=\mathbf{x} \otimes \mathbf{m}$ is the masked image that contains missing pixels, where $\otimes$ is the element-wise multiplication. The masked image $\mathbf{\hat{x}}$ is first fed into the encoder of P-VQVAE to get the patch-based feature vectors. Then UQ-Transformer takes the feature vectors as input and predicts the tokens (\textit{i.e.}, indices) of latent vectors in a codebook for masked regions. Finally, the retrieved latent vectors are fed into the decoder of P-VQVAE to reconstruct the inpainted image, which is guided by the input image from a reference branch.

	\subsection{P-VQVAE}
	\label{sec: p_vqvae}
	P-VQVAE contains a patch-based encoder that maps image patches to feature vectors, a dual-codebook that quantizes feature vectors to discrete tokens, and a multi-scale guided decoder that recovers inpainted images from tokens by referring to the input masked images.

	\label{sec: p_vqvae}
	\subsubsection{Patch-based Encoder}
	\label{sec: p_enc}
	Conventional CNN based encoders process the input image with several convolution kernels in a sliding window manner, which is unsuitable for image inpainting for transformers since the disturbance between masked and unmasked regions is introduced. Thus, the encoder of P-VQVAE (denoted as P-Enc) is designed to process the input image by several linear layers in a non-overlapped patch manner. Specifically, the masked image $\mathbf{\hat{x}}$ is first partitioned into $\frac{H}{r} \times \frac{W}{r}$  non-overlapped patches, where $r$ is the spatial size of patches. For a patch, we call it a \emph{masked patch} if it contains any missing pixels, otherwise an \emph{unmasked patch}. Each patch is flattened and then mapped into a feature vector. Formally, all feature vectors are denoted as $\mathbf{\hat{f}} = \mathcal{E}(\mathbf{\hat{x}})\in \mathbb{R}^{\frac{H}{r}\times\frac{W}{r}\times D}$, where $D$ (set to 256 by default) is the dimension of feature vectors and $\mathcal{E}(\cdot)$ is the encoder function.

	\subsubsection{Dual-Codebook for Vector Quantization}
	\label{sec: d_codes}
	Following previous works~\cite{van2017neural, razavi2019generating, esser2021taming}, the feature vectors from the encoder are quantized into discrete tokens with the latent vectors in the learnable codebook. But differently, we design a dual-codebook (denoted as D-Codes) for vector quantization, which is more suitable for image inpainting. In D-Codes, the latent vectors are divided into two types, denoted as $\mathbf{e} \in \mathbb{R}^{K\times D}$ and $\mathbf{e'} \in \mathbb{R}^{K'\times D}$, which are responsible for feature vectors that are mapped from unmasked and masked patches, respectively. $K$ and $K'$ are the number of latent vectors. Let $\mathbf{m}^{\downarrow} \in [0,1]^{\frac{H}{r}\times \frac{W}{r} \times 1}$ be the unmasked ratio of image patches, \textit{ i.e.}, the ratio of the number of unmasked pixels to $r^2$. It is used as an indicator mask that indicates whether a patch is a masked (with a value less than 1) or unmasked (with a value equal to 1) patch. For each feature vector $\mathbf{\hat{f}}_{i,j}$ in $\mathbf{\hat{f}}$, we use $\mathbf{d}_{i,j} \in \mathbb{R}^{K}$  and $\mathbf{d'}_{i,j} \in \mathbb{R}^{K'}$ as the Euclidean distance between $\mathbf{\hat{f}}_{i,j}$  and the latent vectors in $\mathbf{e}$ and $\mathbf{e'}$. The quantized vector for $\mathbf{\hat{f}}_{i,j}$ is obtained as:
	\begin{equation}
		\begin{cases}
			\mathbf{e}_k \ {\rm where} \ k = {\rm min_{ind}}[\mathbf{d}_{i,j}], \  {\rm if} \   \mathbf{m}^{\downarrow}_{i,j} = 1, \\
			\mathbf{e'}_{k'} \ {\rm where} \ k' = {\rm min_{ind}}[\mathbf{d'}_{i,j}], \ {\rm else}, 
		\end{cases}
		\label{eq: vector_quantization}
	\end{equation}
	where ${\rm min_{ind}}[\cdot]$ is the operation that gets the index of the minimum element in the given vector.
	The quantized vectors and tokens for $\mathbf{\hat{f}}$ are denoted as $\mathbf{\hat{e}} \in \mathbb{R}^{\frac{H}{r} \times \frac{W}{r} \times D}$ and $\mathbf{\hat{t}} = \mathcal{I}(\mathbf{\hat{f}}, \mathbf{e}, \mathbf{e'}, \mathbf{m}^{\downarrow}) \in \mathbb{N}^{\frac{H}{r} \times \frac{W}{r}}$, where $\mathcal{I}(\cdot, \cdot, \cdot, \cdot)$ is the function that gets tokens for its first argument by getting the indices of all quantized vectors in $\mathbf{\hat{e}}$.

	\subsubsection{Multi-Scale Guided Decoder}
	\label{sec: msg_dec}
	For the image inpainting task, an indisputable fact is that the unmasked regions should be kept unchanged. To this end, we design a multi-scale guided decoder, denoted as MSG-Dec, to construct the inpainted image $\mathbf{\hat{x}}^{I}$ by referencing the input masked image $\mathbf{\hat{x}}$. Let $\mathbf{\hat{t}}^{I}$ be the inpainted tokens by transformer (Ref. \Sref{sec: sampling_strategy}) and $\mathbf{\hat{e}}^{I}$ be the retrieved quantized vectors from the codebook based on $\mathbf{\hat{t}}^I$. The construction procedure is formulated as:
	\begin{equation}
		\mathbf{\hat{x}}^{I} = \mathcal{D}(\mathbf{\hat{e}}^{I}, \mathbf{m}, \mathbf{\hat{x}}),
		\label{eq: construction_of_inpainted_image}
	\end{equation}
	where $\mathcal{D}(\cdot, \cdot, \cdot)$ is the decoder function. The decoder consists of two branches: 1) a main branch which starts with the quantized vectors $\mathbf{\hat{e}}^{I}$ and uses several upsampling operations and convolution layers to generate inpainted images. 2) a reference branch that extracts multi-scale feature maps (with spatial sizes $\frac{H}{2^{l}} \times \frac{W}{2^{l}}$, $0 \le l \le {\rm log_2}r$) from the input masked image $\mathbf{\hat{x}}$. The features from the reference branch are fused to the features with the same scale from the main branch through a Mask Guided Addition (MGA) module that is defined below:
	\begin{equation}
		\mathbf{\hat{e}}^{I,l-1} = \mathcal{D}_{l-1}( (1-{\rm int}[\mathbf{m}^{\downarrow,l}]) \otimes \mathbf{\hat{e}}^{I,l} \oplus {\rm int}[\mathbf{m}^{\downarrow,l}] \otimes \mathbf{\hat{f}}^{R,l}),
		\label{eq: mask_guided_addition}
	\end{equation}
	where $\mathbf{\hat{e}}^{I,l}$ and $\mathbf{\hat{f}}^{R,l}$ are the features with spatial size $\frac{H}{2^{l}} \times \frac{W}{2^{l}}$ from the main branch and the reference branch, respectively. $\mathbf{m}^{\downarrow,l}$ is the unmasked ratio obtained from $\mathbf{m}$ for the corresponding spatial size. $\oplus$ is the elementwise addition operation. $\mathcal{D}_{l-1}(\cdot)$ is the sub-network in the main branch that consists of an upsampling operation, a convolution layer, and several Conv ResBlocks. By removing the reference branch in the decoder, P-VQVAE is also capable of reconstructing the input image $\mathbf{\hat{x}}$ like other auto-encoders.

	\begin{figure}[t]
		\centering
		\includegraphics[width=1.0\columnwidth]{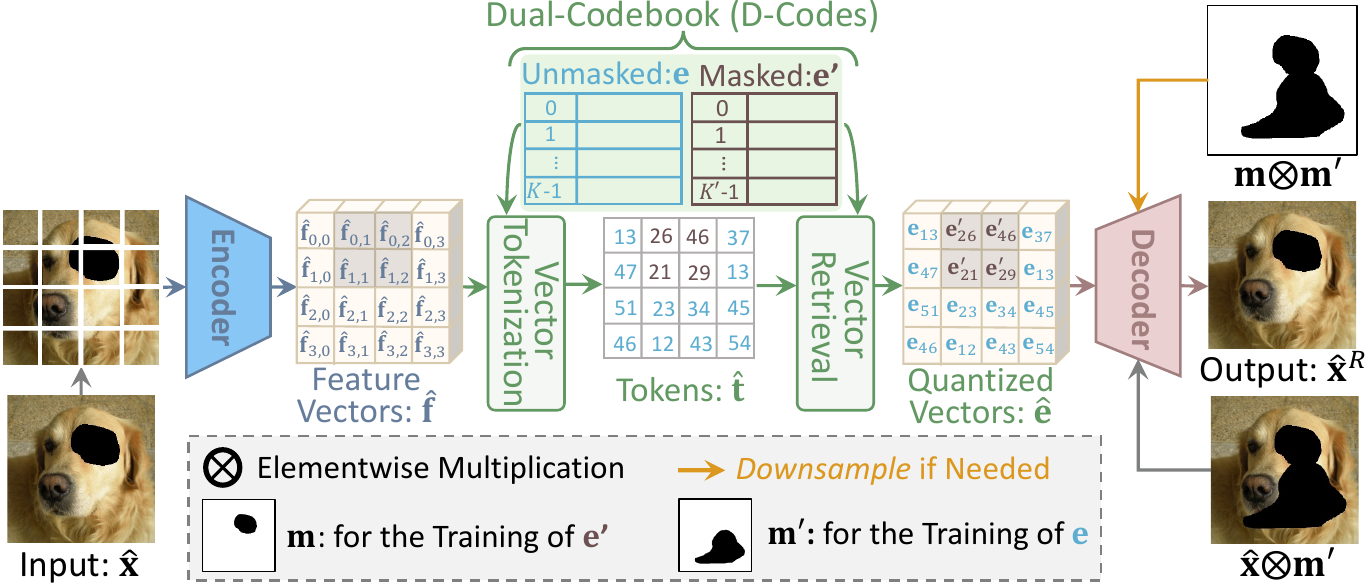} 
		\caption{Training procedure of P-VQVAE. The unmasked pixels in the reference image $\mathbf{\hat{x}} \otimes \mathbf{m'}$ are utilized to recover the corresponding pixels in $\mathbf{\hat{x}}^R$, while the latent vectors in codebook $\mathbf{e}'$ and $\mathbf{e}$ are used to recover the pixels in $\mathbf{\hat{x}}^R$ masked by $\mathbf{m}$ and the remaining pixels, respectively.}
		\label{figure: p_vqvae_framework}
	\end{figure}

	\subsubsection{Training of P-VQVAE}
	The decoder of P-VQVAE learns to reconstruct the inpainted image based on not only the retrieved quantized vectors but also the input masked image from the reference branch. Therefore, to avoid the decoder learning to reconstruct the input image $\mathbf{\hat{x}}$ only from the reference  image, we get the reference image by randomly erasing some pixels in $\mathbf{\hat{x}}$  with another mask $\mathbf{m'}$ (see in \Fref{figure: p_vqvae_framework}). Let $\mathbf{\hat{x}}^R = \mathcal{D}(\mathbf{\hat{e}}, \mathbf{m} \otimes \mathbf{m'}, \mathbf{\hat{x}} \otimes \mathbf{m'})$ be the reconstructed image. In our design, the unmasked pixels in the reference image are utilized to recover the corresponding pixels in $\mathbf{\hat{x}}^R$, while the latent vectors in codebook $\mathbf{e}'$ and $\mathbf{e}$ are used to recover the pixels in $\mathbf{\hat{x}}^R$ masked by $\mathbf{m}$ and the remaining pixels, respectively. P-VQVAE is trained with the commonly used VQ-VAE loss~\cite{van2017neural}. Please refer to our conference paper~\cite{liu2022reduce} for more details.

	\subsubsection{Learning More Latent Vectors}	
	Different from the settings in our conference paper, more latent vectors in the codebook are learned in this work to help the model distinguish more patterns and textures. However, when the number of latent vectors in codebook $\mathbf{e}$ is increased, for example, from 512 to 8192, we empirically find that only a few of them are used (about $3000$). To handle this, we learn the latent vectors with Gumbel-Softmax relaxation~\cite{jang2016categorical} in the training stage. The core idea in Gumbel-Softmax relaxation is to add some noise to the distance between feature vectors and latent vectors, so as to avoid the phenomenon that only a subset of latent vectors are used. Specifically, the quantized vector for feature vector $\mathbf{\hat{f}}_{i,j}$ obtained in \Eref{eq: vector_quantization} is rewritten as:
	\begin{equation}
		\begin{cases}
			\mathbf{e}_k \ {\rm where} \ k = {\rm min}_{\rm ind} [-\mathcal{S}(\mathbf{n} \ominus \mathbf{d}_{i,j}, \tau)], \  {\rm if} \   \mathbf{m}^{\downarrow}_{i,j} = 1 \\
			\mathbf{e'}_{k'} \ {\rm where} \ k' = {\rm min}_{\rm ind} [-\mathcal{S}(\mathbf{n'} \ominus \mathbf{d'}_{i,j}, \tau)], \ {\rm else}, 
		\end{cases}
		\label{eq: vector_quantization_with_gumbel}
	\end{equation}
	where $\mathbf{n} \in \mathbb{R}^{K}$ and $\mathbf{n'} \in \mathbb{R}^{K'}$ are the sampled Gumbel noises, $\mathcal{S}(\cdot, \tau)$ is the softmax function applied to its first argument with temperature $\tau$. More training details are provided in \Sref{sec: implementation_details}.

	\subsection{UQ-Transformer}
	\label{sec: uq_transformer}
	In existing transformers for image inpainting~\cite{wan2021high} and synthesis~\cite{esser2021taming, ramesh2021zero}, the quantized discrete tokens are used as both the inputs and prediction targets. Given such discrete tokens, transformers suffer from the severe input information loss issue, which is harmful to their prediction. In contrast, to take full advantage of feature vectors $\mathbf{\hat{f}}$ from the encoder of P-VQVAE, our UQ-Transformer takes $\mathbf{\hat{f}}$ rather than the discrete tokens as inputs and predicts the discrete tokens for masked patches.

	\subsubsection{Mask Embedding}
	\label{sec: mask_embeding}
	As a common practice, a set of position embeddings $\mathbf{f}^P \in \mathbb{R}^{\frac{H}{r} \times \frac{W}{r}\times D'}$ is learned to introduce positional induction in transformers, where $D'$ is the hidden dimensionality before UQ-Transformer. However, we find that UQ-Transformer cannot distinguish the patch embeddings for masked and unmasked patches effectively. \qiankun{The reasons come from two folds: 1) The numbers of missing pixels in different masked patches are not restricted to be the same. 2) The missing pixels in a masked image are set to zeros, which are the same as unmasked \textit{black} pixels.} This makes PUT produce some artifacts with \emph{black} pixels in masked regions easily when the image contains some natural black pixels, which will be analyzed in the following \Sref{sec:effectiveness_of_mask_embedding}. To alleviate this issue, we introduce a simple yet effective mask embedding $\mathbf{f}^M \in \mathbb{R}^{1 \times 1 \times D'}$. \qiankun{Similar to the position embeddings $\mathbf{f}^P$, the mask embedding $\mathbf{f}^M$ is also learned during the training stage. But differently, the image patches in different locations share the same mask embedding.} Let $\mathbf{\bar{f}} \in \mathbb{R}^{\frac{H}{r} \times \frac{W}{r}\times D'}$ be the input features of the first transformer block, it can be obtained as:
	\begin{equation}
		\mathbf{\bar{f}} = (\mathbf{m}^{\downarrow} \otimes   \mathcal{F}(\mathbf{\hat{f}}) \oplus (1 - \mathbf{m}^{\downarrow}) \otimes \mathbf{f}^M) \oplus \mathbf{f}^P, 
		\label{eq: feaures_feeding_transformer_block}
	\end{equation}
	where $\mathcal{F}(\cdot)$ is a linear layer that adjusts the dimensionality of the input features. \qiankun{The features of image patches are added with mask embedding according to their mask ratios $\mathbf{m}^{\downarrow}$, which helps the model distinguish the unmasked and masked patches, as well as the masked patches that with different numbers of missing pixels.}

	\subsubsection{Prediction of UQ-Transformer}
	The output of the last transformer block is projected to the distribution over $K$ latent vectors in codebook $\mathbf{e}$ with a linear layer and a softmax function. We formulate the above procedure as $\mathbf{\hat{p}} = \mathcal{T}(\mathbf{\hat{f}}) \in [0, 1]^{\frac{H}{r}\times \frac{W}{r} \times K}$, where $\mathcal{T}(\cdot)$ refers to the UQ-Transformer function. With the predicted probability, the tokens for masked patches can be sampled.

	The intuition behind this design is that UQ-Transformer learns the likelihood of the tokens for masked patches given the unmasked patch embeddings. Let $\Pi=\{(i,j)|\mathbf{m}^{\downarrow}_{i,j}<1\}$ be the index set of masked patches. For each masked patch with index $(i,j)$, the likelihood can be denoted as $p(\mathbf{\hat{t}}^{I}_{i,j}|\mathbf{\hat{f}}_{\bar{\Pi}}) = \mathbf{\hat{p}}_{i,j}$, where $\bar{\Pi}$ is the complementary set of $\Pi$, \emph{i.e.}, the index set of unmasked patches.

	\subsubsection{Training of UQ-Transformer}
Given a masked image $\mathbf{\hat{x}}$, the distribution of its corresponding inpainted tokens over $K$ latent vectors can be obtained with the pre-trained P-VQVAE and UQ-Transformer $\mathbf{\hat{p}} = \mathcal{T}(\mathcal{E}(\mathbf{\hat{x}}))$.
The ground-truth tokens for $\mathbf{x}$ are $\mathbf{t} = \mathcal{I}(\mathcal{E}(\mathbf{x}), \mathbf{e}, \mathbf{e'}, \mathcal{O}(\mathbf{m}^{\downarrow}))$  (Ref. \Sref{sec: d_codes}), where $\mathcal{O}(\cdot)$ sets all values in the given argument to 1. 
UQ-Transformer is trained with a classification loss by fixing P-VQVAE:
\begin{equation}
    \begin{aligned}
        L_{trans} = -\frac{1}{|\Pi|} \sum_{(i,j) \in \Pi} {\rm log}(\mathbf{\hat{p}}_{i,j,\mathbf{t}_{i,j}}).
    \end{aligned}
    \label{eq: uq_transformer_loss}
\end{equation}
In order to make the training stage consistent with the inference stage, where only the quantized vectors can be obtained for masked regions, we randomly quantize the feature vectors in $\mathcal{E}(\mathbf{\hat{x}})$ to the latent vectors in the codebook with probability 0.3 before feeding it to UQ-Transformer.

	\subsection{Sampling Strategy for Image Inpaining}
	\label{sec: sampling_strategy}
In order to get diverse results, existing transformer based works usually adopt a per-token sampling strategy for image synthesis~\cite{radford2019language, ramesh2021zero, esser2021taming} and inpainting~\cite{wan2021high}, \emph{i.e.}, the tokens are sampled one-by-one. Formally, let $\bar{\Pi}^I$ be the index set of previously inpainted patches and $\mathbf{\hat{e}}^{I}_{\bar{\Pi}^I}$ be the retrieved latent vectors for inpainted patches. The likelihood of masked patch $(i,j)$ is obtained given the unmasked patches embeddings $\mathbf{\hat{f}}_{\bar{\Pi}}$ and the retrieved latent vectors $\mathbf{\hat{e}}^{I}_{\bar{\Pi}^I}$:
\begin{equation}
    p(\mathbf{\hat{t}}^{I}_{i,j}|\mathbf{\hat{f}}_{\bar{\Pi}}, \mathbf{\hat{e}}^I_{\bar{\Pi}^I}) = \mathcal{T}(\mathcal{R}(\mathbf{\hat{f}}, \mathbf{\hat{e}}^{I}_{\bar{\Pi}^I}))_{i,j},
\end{equation}
where $\mathcal{R}(\mathbf{\hat{f}}, \mathbf{\hat{e}}^{I}_{\bar{\Pi}^I})$ replaces the feature vectors of masked patches in $\mathbf{\hat{f}}$ with retrieved latent vectors $\mathbf{\hat{e}}^{I}_{\bar{\Pi}^I}$ according to the indices in $\bar{\Pi}^I$. \qiankun{$|\Pi|$ iterations are required to finish the inpainting process, which is time-consuming.}

To better balance the inference time and image quality, we design a multi-token sampling strategy. \qiankun{The overall multi-token sampling strategy is divided into two steps: 1) selecting the patches to be inpainted, and 2) sampling the tokens for selected patches.} Specifically, in each iteration, we first get an index set, denoted as $\Pi'$, of masked patches with top-$\mathcal{K}_1$ maximum predicted probabilities among the remaining masked patches:
\begin{equation}
    \Pi' = {\rm TopK}_{\rm ind}[{\rm max}[ \mathcal{T}(\mathcal{R}(\mathbf{\hat{f}}, \mathbf{\hat{e}}^{I}_{\bar{\Pi}^I}))]_{\Pi - \bar{\Pi}^{I}}, \mathcal{K}_1],
\end{equation}
where ${\rm max}[\cdot]$ returns the maximum values of its argument along the last dimensionality, $\Pi - \bar{\Pi}^{I}$ is the indices of remaining masked patches and ${\rm TopK_{\rm ind}[\cdot, \mathcal{K}_1]}$ gets the indices of the top-$\mathcal{K}_1$ maximum values in its first argument. Then the tokens for these $\mathcal{K}_1$ patches are simultaneously but independently sampled from the likelihood:
\begin{equation}
    p(\mathbf{\hat{t}}^{I}_{\Pi '}|\mathbf{\hat{f}}_{\bar{\Pi}}, \mathbf{\hat{e}}^I_{\bar{\Pi}^I}) = \mathcal{T}(\mathcal{R}(\mathbf{\hat{f}}, \mathbf{\hat{e}}^{I}_{\bar{\Pi}^I}))_{\Pi '}.
\end{equation}

	\begin{figure}[t]
		\centering
		\includegraphics[width=1.0\columnwidth]{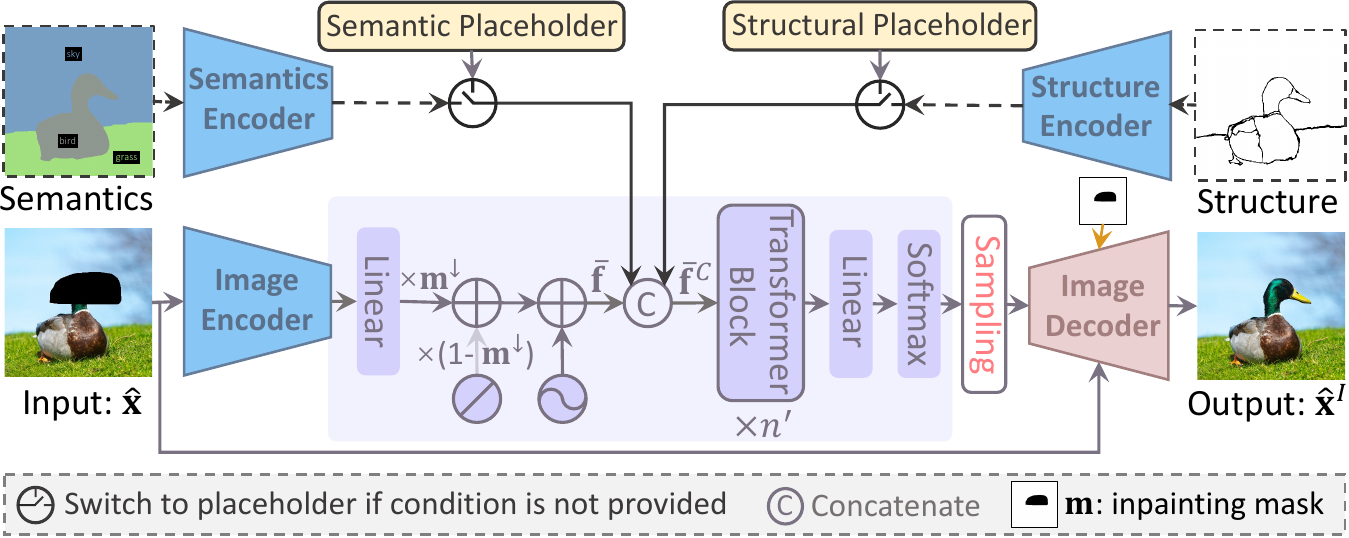} 
		\mycaption{The pipeline of controllable image inpainting. Users are free to provide none, one, or both of these two conditions.}
		\label{figure: controllable_inpainting}
	\end{figure}

	To avoid sampling the tokens that have low probabilities, we only maintain the top-$\mathcal{K}_2$ elements in the likelihood for each patch while sampling. \qiankun{The two parameters $\mathcal{K}_1$ and $\mathcal{K}_2$ mainly have impacts on inference speed and image quality, respectively. However, as we will see in \Sref{sec: dicussion_on_sampling_strategy}, PUT is robust to different values of them to some extent. A special case of our multi-token sampling strategy is that $\mathcal{K}_1 = |\Pi|$, which means the tokens of all masked patches are simultaneously sampled in one iteration.}

	\qiankun{\subsection{Controllable Image Inpainting}}
\qiankun{Though PUT generates photorealistic and diverse results for a masked image, it is hard for end-users to intervene in the sampling process to control the generated contents for masked regions. In this section, we introduce semantic and structural conditions (\textit{i.e.}, segmentation and sketch maps) provided by users as extra guidance for PUT, making the inpainted results controllable, as shown in \Fref{figure: controllable_inpainting}.}

\qiankun{\subsubsection{Semantic and Structural Maps as Features}}
\qiankun{Semantic and structural maps are intuitive to be drawn and understood by users. However, there are two challenges for the drawing of semantic maps: 1) the categories in real-world applications are open-set, while most existing high-quality segmentation models are trained on closed-set categories. 2) not all users are professional enough to categorize the content in an image and relate it to the supported categories accurately. To handle these challenges, we design a simple unknown category strategy that allows the user to label categories not included in the closed-set categories. Since an image may contain multiple unrecognized objects with different categories, multiple unknown categories are needed in the model. Therefore, we define a certain number of unknown categories. During the training procedure, we randomly select some of the known categories in the semantic map and use the same number of unknown categories to substitute them. During inference, if users encounter objects with categories not included in the closed-set categories, as shown in \Fref{figure: unknown_category}, the user can manually label these objects using multiple unknown categories.}

\qiankun{Formally, we denote the user-provided semantic and structural maps with $\mathbf{c}_{sem} \in \{0,1,..., C+U-1\}^{H\times W \times 1}$ and $\mathbf{c}_{str} \in \{0,1\}^{H\times W \times 1}$, where $C$ is the number of known categories while $U$ is the pre-defined number of unknown categories. Since the content of masked regions is predicted by UQ-Transformer in the feature domain, we need to map semantic and structural maps into feature representation as well. To this end, two P-VQVAEs (without reference branch) are trained to extract features from these two types of maps. The extracted features for semantic and structural maps are denoted as $\mathbf{f}_{sem}, \mathbf{f}_{str} \in \mathbb{R}^{\frac{H}{r}\times \frac{W}{r} \times D}$, respectively.}

\qiankun{\subsubsection{Semantic and Structural Features as Conditions}} 
\qiankun{Once the semantic and structural features are obtained from condition maps, we concatenate them with the features obtained by \Eref{eq: feaures_feeding_transformer_block} along the last dimension:
\begin{equation}
	\mathbf{\bar{f}}^{C} = {\rm Concat}(\mathbf{\bar{f}}, \mathbf{f}_{sem}, \mathbf{f}_{str}).
	\label{eq:input_emb_with_two_conditions}
\end{equation}
The feature $\mathbf{\bar{f}}^{C} \in \mathbb{R}^{\frac{H}{r}\times \frac{W}{r} \times (D'+2D)}$ is used to replace $\mathbf{\bar{f}}$ as the input of the first transformer block.}

\qiankun{Though users can achieve controllability over inpainted images with the above two types of conditions, it is not flexible enough in the inference stage if users are always asked to provide these two types of conditions. Therefore, we introduce a placeholder embedding for each type of condition, which has the shape of $1 \times 1 \times D$ and is learned during the training stage. When a certain type of condition is absent, the corresponding placeholder embedding is repeated and expanded to the shape $\frac{H}{r}\times \frac{W}{r} \times D$ and used as the features to be concatenated in \Eref{eq:input_emb_with_two_conditions}. With the help of placeholder embeddings, users are allowed to provide none, one, or both of these two types of conditions, making PUT more flexible in real-world applications.}

\qiankun{In the training stage of UQ-Transformer, the classification loss in \Eref{eq: uq_transformer_loss} is used and the condition features are replaced with their corresponding placeholder embeddings randomly with a probability of 0.3.}
		
	\begin{figure*}[t]
		\centering
		\includegraphics[width=2.05\columnwidth]{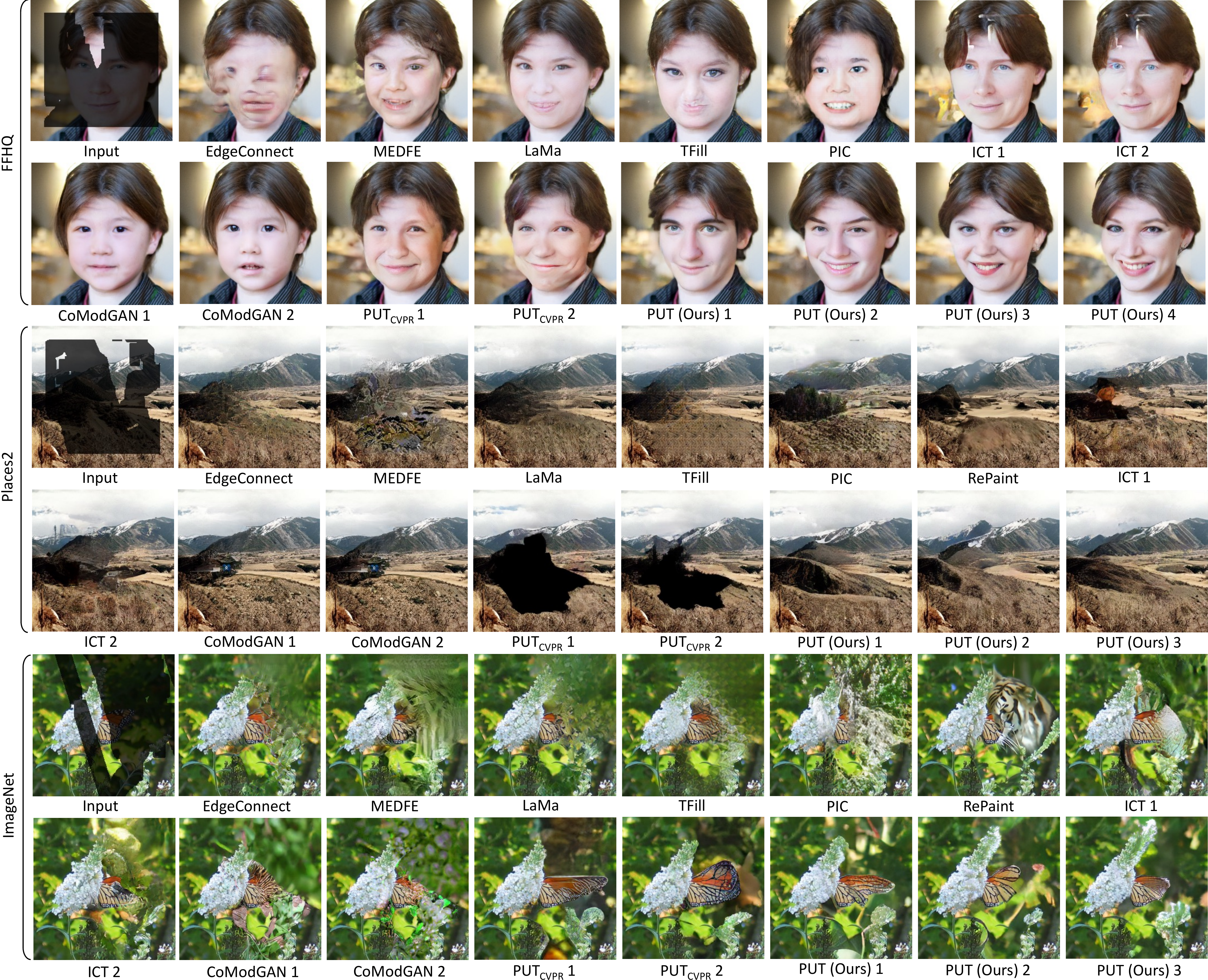} 
		\caption{Inpainting results produced by different methods on different datasets. The images produced by PUT are of higher diversity and visual quality.}
		\label{figure: inpainting_results_256}
	\end{figure*}

	\section{Experiments}
	\label{sec: experiments}
	\subsection{Datasets}
	The evaluation is conducted on FFHQ~\cite{karras2019style}, Places2~\cite{zhou2017places}, and ImageNet~\cite{deng2009imagenet}. We use the original training and testing splits for Places2 and ImageNet. For FFHQ, we maintain the last 1K images for evaluation and use the others for training. Following ICT~\cite{wan2021high}, 1K images are randomly chosen from the test split of ImageNet for evaluation, and the irregular masks provided by PConv~\cite{liu2018image} are used for both training and testing. \qiankun{While applying the pretrained transformer in PUT to downstream tasks, COCO~\cite{lin2014microsoft} and LVIS~\cite{gupta2019lvis} are used for the evaluation of object detection and segmentation.}

	\begin{figure}[t]
			\centering
			\includegraphics[width=1.0\columnwidth]{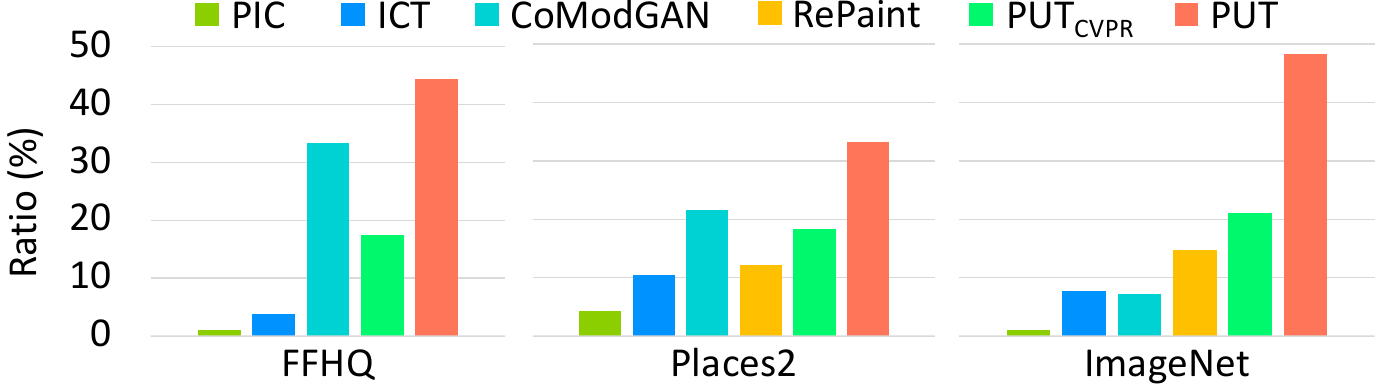} 
			\caption{The ratio of each method that is chosen as the rank 1st result by humans. Statistics are collected from 20 participants. Note that the diffusion model based  RePaint~\cite{lugmayr2022repaint} is not involved in the evaluation on FFHQ~\cite{karras2019style} since there is no available model. Reported on $256 \times 256$ resolution.}
			\label{figure: usr_study}
		\end{figure}

	\subsection{Implementation Details}
	\label{sec: implementation_details}
We set $r=8$ and $r=16$ for resolution $256 \times 256$ and $512 \times 512$, respectively, making the encoder produce features that have the same spatial size $32 \times 32$. To change the resolutions of features in the decoder, we adopt the nearest upsampling/downsampling. We train P-VQVAE with Adam~\cite{kingma2014adam}, where the coefficients used for computing running averages of gradient and its square are set to 0 and 0.9. The batch size is set to 192 for FFHQ and Places2 and 384 for ImageNet. The learning rate is warmed up from 0 to 2e-4 linearly in the first 5K iterations and then decayed with a cosine scheduler. For different datasets, P-VQVAEs are all optimized for 100 epochs. In our conference paper~\cite{liu2022reduce}, the number of latent vectors is set to $K=512$ and $K'=512$, which are increased to $K=8192$ and $K'=1024$  in this paper. The Gumbel-Softmax relaxation~\cite{jang2016categorical} is adopted during the whole training procedure. Specifically, the Gumbel relaxation temperature is annealed from 20 to 1e-6 over the first 5K iterations using a cosine annealing and the scale of added Gumbel noise is dropped from $1$ to $0.1$ after the 5K-th iteration.

	The UQ-Transformer in this paper is modified from ViT-Base~\cite{dosovitskiy2020image}: the convolution layer for patch embedding and the classification token for image classification are removed. For different datasets, we use UQ-Transformer with the same number of parameters, which differs from the settings in our conference paper~\cite{liu2022reduce} (see \Tref{table:flops_and_params}). Given the pretrained P-VQVAE, UQ-Transformer is optimized with AdamW~\cite{loshchilov2017decoupled}. The coefficients for computing running average and square of gradient are set to 0.9 and 0.95. The batch size is 192 for FFHQ and Places2 and 384 for ImageNet. The learning rate is increased from 1e-5 to 1.5e-3 linearly during the first 20K iterations. All models are trained to their convergence. \qiankun{We set $D'$ to 256/768 for PUT with/without conditions to ensure that the hidden dimension of UQ-Transformer is the same as ViT-Base. For controllable image inpainting, we use the segmentation map predicted by Mask2Former~\cite{cheng2021mask2former} (with $C=133$ known categories) and the sketch map predicted by DexiNed~\cite{soria2023dexined_ext} to train the model. The number of unknown categories $U$ is set to 20.}

	We compare PUT with several state-of-the-art methods. The method in our conference paper~\cite{liu2022reduce} is denoted as ``PUT$_{\rm CVPR}$". For a fair comparison, we directly use the pre-trained models provided by the authors when available, otherwise, we train the models ourselves using the official codes and settings. \qiankun{Since existing inpainting methods do not support controllable inpainting, we compare PUT with other methods without using conditions. The controllability of PUT is discussed and evaluated in \Sref{sec: controllable inpainting}.	}

		\begin{table*}[t]
			\setlength{\tabcolsep}{7.5pt}
			\centering
			\caption{Quantitative results for resolution $256 \times 256$.
				PUT samples the tokens for all masked patches within one iteration for the comparison with deterministic methods (top group), but samples the tokens for 20 masked patches iteratively for the comparison with pluralistic methods (bottom group).}
			\label{table:quantitative_results_256}
			\begin{tabular}{c|c|ccc|ccc|ccc}
				\hline
				\multicolumn{2}{c|}{Dataset} & \multicolumn{3}{c|}{FFHQ~\cite{karras2019style}} & \multicolumn{3}{c|}{Places2~\cite{zhou2017places}} & \multicolumn{3}{c}{ImageNet~\cite{deng2009imagenet}} \\ 
				\hline
				\multicolumn{2}{c|}{Mask Ratio (\%)} & 20-40 & 40-60 & 10-60 & 20-40 &40-60 & 10-60 & 20-40 & 40-60 & 10-60 \\
				\hline
				\multirow{13}*{FID $\downarrow$} 
				& EdgeConnect (ICCVW, 2019)~\cite{nazeri2019edgeconnect}  & 12.949 & 26.217 & 16.961 & 20.180 &34.965 & 23.206 & 27.821 & 63.768 & 39.199\\
				& MEDFE (ECCV, 2020)~\cite{liu2020rethinking} &13.999 & 26.252 & 17.061 & 28.671 &46.815 &32.494 &40.643 &93.983 &54.854 \\
				& LaMa (WACV, 2022)~\cite{suvorov2022resolution} &\textbf{\underline{9.757}} &\textbf{\underline{18.648}} &13.786 &20.268 &33.216 &\qiankun{22.599} &18.988 &45.051 &26.854 \\
				&TFill (CVPR, 2022)~\cite{zheng2022bridging} &11.340 &19.486 &12.984 &19.512 &34.022 &23.753 &24.739 &50.670 &32.342 \\ 
				& ICT  (ICCV, 2021)~\cite{wan2021high} &10.442 &23.946 &15.363 &19.309 &33.510 &23.331 &23.889 &54.327 &32.624 \\
				& PUT$_{\rm CVPR}$ (CVPR, 2022)~\cite{liu2022reduce} &11.221 &19.934 &13.248 &19.776 &38.206 &24.605 & 19.411 &43.239 &26.223\\
				& PUT (Ours) &10.844 &18.842 &\textbf{\underline{12.728}} &\underline{\bf 18.219} &\underline{\bf29.758} &\qiankun{\underline{\bf 21.234}}&\textbf{\underline{18.411}} &\textbf{\underline{41.441}} &\textbf{\underline{24.794}}  \\
				\cline{2-11}
				& PIC (CVPR, 2019)~\cite{zheng2019pluralistic} & 22.847 & 37.762 & 25.902 & 31.361 & 44.289 & 34.520 & 49.215 & 102.561 & 63.955 \\
				& CoModGAN (ICLR, 2021)~\cite{zhao2021comodgan} &\underline{\bf 10.102} &\underline{\bf 17.770} &\underline{\bf 12.236} &20.804 &33.931 &25.367 &29.414 &64.020 &38.652 \\
				& ICT (ICCV, 2021)~\cite{wan2021high} &13.536 &23.756 &16.202 &20.900 & 33.696 &24.138 &25.235 &55.598 &34.247  \\ 
				&RePaint (CVPR, 2022)~\cite{lugmayr2022repaint} &- &- &- &\underline{\bf 17.792} & 31.934 &21.732 &24.525 &60.476 &35.723 \\
				& PUT$_{\rm  CVPR}$ (CVPR, 2022)~\cite{liu2022reduce} & 12.784 &21.382 &14.554 &19.617 &31.485 & 22.121 &21.272 &45.153 &27.648\\
				& PUT (Ours) &11.891 &19.458 &13.805 &19.028 &\underline{\bf 29.122} &\underline{\bf 21.158} &\underline{\bf 19.956} &\underline{\bf 40.329} &\underline{\bf 25.642} \\
				\hline
				
				\multirow{13}*{LPIPS$\downarrow$} 
				& EdgeConnect (ICCVW, 2019)~\cite{nazeri2019edgeconnect} &0.0974 &0.2055 &0.1330 &\qiankun{0.1217} & 0.2220 &\qiankun{0.1470} &0.1417 &0.2531 &0.1815 \\
				& MEDFE (ECCV, 2020)~\cite{liu2020rethinking} &0.1080 &0.1997 &0.1355 &0.1617 &0.2760 &0.1953 & 0.1826 &0.3169 &0.2219\\
				& LaMa (WACV, 2022)~\cite{suvorov2022resolution} &\textbf{\underline{0.0743}} &0.1716 &0.1104 &0.1330 &0.2260 &0.1496 &0.1147 &0.2205 &0.1463 \\
				&TFill (CVPR, 2022)~\cite{zheng2022bridging} &0.0904 &0.1665 &0.1331 &\qiankun{\underline{\bf 0.1172}} &\qiankun{\underline{\bf 0.2036}}&\qiankun{\underline{\bf 0.1434}} &0.1292 &0.2304 &0.1601  \\ 
				& ICT (ICCV, 2021)~\cite{wan2021high} &0.0814 &0.1838 &0.1219 &0.1224 &0.2231 &0.1534 &0.1263 &0.2425 &0.1636 \\
				& PUT$_{\rm CVPR}$ (CVPR, 2022)~\cite{liu2022reduce} &0.0883 &0.1647 &0.1112 &0.1224 &0.2280 &0.1555 &0.1159 &0.2257 &0.1518 \\
				& PUT (Ours) &0.0870 &\textbf{\underline{0.1605}} &\textbf{\underline{0.1089}} &0.1223 &\qiankun{0.2159} &0.1505 &\textbf{\underline{0.1102}} &\textbf{\underline{0.2163}} &\textbf{\underline{0.1454}} \\
				\cline{2-11}
				&PIC (CVPR, 2019)~\cite{zheng2019pluralistic} &0.1497 &0.2660 &0.1816 &0.1889 &0.2962 &0.2129 &0.2114 &0.3510 &0.2538 \\
				& CoModGAN (ICLR, 2021)~\cite{zhao2021comodgan}  &\underline{\bf 0.0867} &\underline{\bf 0.1705} &\underline{\bf 0.1126} &\underline{\bf 0.1177} &\underline{\bf 0.2170} &\underline{\bf 0.1416} &0.1479 &0.2717 &0.1857 \\
				& ICT (ICCV, 2021)~\cite{wan2021high} &0.1001 &0.1944 &0.1276 &0.1285 &0.2306 &0.1535 &0.1348 &0.2567 &0.1752 \\
				&RePaint (CVPR, 2022)~\cite{lugmayr2022repaint} &- &- &- &0.1265 &0.2394 &0.1628 &0.1310 &0.2673 &0.1767 \\
				& PUT$_{\rm CVPR}$ (CVPR, 2022)~\cite{liu2022reduce} &0.0983 &0.1818 &0.1231 &0.1265 &0.2255 &0.1569 &0.1263 &0.2409 &0.1616 \\
				& PUT (Ours) &0.0941 &0.1712 &0.1167 &0.1252 &0.2193 &0.1532 &\underline{\bf 0.1184} &\underline{\bf 0.2256} &\underline{\bf 0.1527} \\
				\hline

				\multirow{13}*{PSNR $\uparrow$} 
				& EdgeConnect (ICCVW, 2019)~\cite{nazeri2019edgeconnect} & 27.484 & 22.574 & 26.181 & 26.536 &22.755 &25.975 & 24.703 &20.459 &23.596\\
				& MEDFE (ECCV, 2020)~\cite{liu2020rethinking} &27.117 & 22.499 & 26.111 & 25.401 &21.543 &24.510 &23.730 &19.560 &22.752\\
				& LaMa (WACV, 2022)~\cite{suvorov2022resolution} &29.834 &\textbf{\underline{24.463}} &27.545 &\textbf{\underline{27.004}} &\textbf{\underline{23.260}} &\textbf{\underline{26.803}} & \textbf{\underline{26.001}} & \textbf{\underline{21.614}} & \textbf{\underline{25.140}} \\
				&TFill (CVPR, 2022)~\cite{zheng2022bridging} &27.171 &22.915 &26.260 &26.699 &23.177 &25.916 &24.855 &20.970 &23.965\\ 
				& ICT (ICCV, 2021)~\cite{wan2021high} &\textbf{\underline{29.847}} &23.041 &26.736 &25.836 &22.120 &24.986 &24.249 &20.045 &23.317 \\
				& PUT$_{\rm CVPR}$ (CVPR, 2022)~\cite{liu2022reduce} &28.356 &24.125 &27.473 &26.580 &22.945 &25.749 &25.721 &21.551 &24.726 \\
				& PUT (Ours) &28.538 &24.200 &\textbf{\underline{27.616}} &26.220 &22.579 &25.442 &25.736 &21.504 &24.719 \\
				\cline{2-11}
				&PIC (CVPR, 2019)~\cite{zheng2019pluralistic} & 25.157 & 20.424 & 24.093 & 24.073 & 20.656 & 23.469 & 22.921 & 18.368 & 21.623\\
				& CoModGAN (ICLR, 2021)~\cite{zhao2021comodgan}  &\qiankun{27.261} &22.411 &26.165 &\underline{\bf 26.333} & 21.986 &\underline{\bf 25.120} &23.882 &19.412 &\qiankun{\underline{\bf 23.839}} \\
				& ICT (ICCV, 2021)~\cite{wan2021high} &26.462 &21.816 &25.515 &24.947 &21.126 &24.373 &23.252 &19.025 &22.123\\
				&RePaint (CVPR, 2022)~\cite{lugmayr2022repaint} &- &- &- &25.146 &20.899 &24.136 &23.268 &18.380 &22.067 \\
				& PUT$_{\rm CVPR}$ (CVPR, 2022)~\cite{liu2022reduce} &26.877 &22.375 &25.943 &25.452 &21.528 &24.492 &24.238 &19.742 &23.264\\
				& PUT (Ours) &\qiankun{\underline{\bf 27.504}} &\underline{\bf 23.127} &\underline{\bf 26.582} &25.757 &\underline{\bf22.028} &24.901 &\underline{\bf 24.542 }&\underline{\bf 20.044} &\qiankun{23.459} \\
				\hline
				
				\multirow{13}*{SSIM$\uparrow$} 
				& EdgeConnect (ICCVW, 2019)~\cite{nazeri2019edgeconnect} & 0.941 & 0.826 & 0.899 &0.881 & 0.734 & 0.840 & 0.882 & 0.714 &0.824 \\
				& MEDFE (ECCV, 2020)~\cite{liu2020rethinking} &0.936 &0.840 &0.903 &0.854 &0.685 &0.796 &0.861 &0.675 &0.795 \\
				& LaMa (WACV, 2022)~\cite{suvorov2022resolution} &0.963 &0.887 &0.923 &\textbf{\underline{0.887}} &0.750 &\textbf{\underline{0.852}} &\textbf{\underline{0.906}} &0.759 &\textbf{\underline{0.857}} \\
				&TFill (CVPR, 2022)~\cite{zheng2022bridging} &0.938 &0.848 &0.907 &\underline{\bf 0.887} &\qiankun{\underline{\bf 0.757}} &0.841 &0.888 &0.741 &0.836\\ 
				& ICT (ICCV, 2021)~\cite{wan2021high} &\textbf{\underline{0.964}} &0.863 &0.917 &0.870 &0.723 &0.819 &0.876 &0.711 &0.818\\
				& PUT$_{\rm CVPR}$ (CVPR, 2022)~\cite{liu2022reduce} &0.953 &0.888 &0.908 &0.885 &\qiankun{0.756} &0.840 &0.904 &0.772 &0.838 \\
				& PUT (Ours) &0.954 &\textbf{\underline{0.890}} &\textbf{\underline{0.933}} &0.877 &0.740 &0.831 &0.904 &\textbf{\underline{0.773}} &0.856 \\
				\cline{2-11}
				&PIC (CVPR, 2019)~\cite{zheng2019pluralistic} & 0.910 & 0.769 & 0.865 & 0.824 & 0.648 & 0.775 & 0.842 & 0.623 & 0.766\\
				& CoModGAN (ICLR, 2021)~\cite{zhao2021comodgan}  &0.939 &0.840 &0.904 &\underline{\bf 0.875} &0.713 &0.811 &0.866 &0.685 &0.803 \\
				& ICT (ICCV, 2021)~\cite{wan2021high} &0.931 &0.822 &0.896 &0.850 &0.682 &0.803 &0.852 &0.666 &0.786 \\
				&RePaint (CVPR, 2022)~\cite{lugmayr2022repaint} &- &- &- &0.853 &0.674 &0.792 &0.852 &0.638 &0.775\\
				& PUT$_{\rm CVPR}$ (CVPR, 2022)~\cite{liu2022reduce} &0.936 &0.845 &0.906 &0.861 &0.703 &0.806 &0.875 &0.704 &0.818\\
				& PUT (Ours) &\underline{\bf 0.944} &\underline{\bf 0.863} &\underline{\bf 0.916} &0.869 &\underline{\bf 0.722} &\underline{\bf 0.818} &\underline{\bf 0.882} &\underline{\bf 0.721} &\underline{\bf 0.826} \\
				\hline
				
			\end{tabular}
			\label{table: comparison_with_d_methods_256}
		\end{table*}

		\subsection{Evaluation on $\mathbf{256 \times 256}$ Resolution} 
		The evaluated methods are EdgeConnect~\cite{nazeri2019edgeconnect}, MEDFE~\cite{liu2020rethinking}, TFill~\cite{zheng2022bridging}, PIC~\cite{zheng2019pluralistic}, ICT~\cite{wan2021high}, RePaint~\cite{lugmayr2022repaint} and PUT$_{\rm CVPR}$~\cite{liu2022reduce}. Among these methods, the first three are deterministic methods, while the last four are pluralistic methods. Some recently proposed methods, including LaMa~\cite{suvorov2022resolution} (deterministic) and CoModGAN~\cite{zhao2021comodgan} (pluralistic), which are mainly designed for $512\times 512$ resolution, are also modified to $256\times 256$ resolution for more comparisons.

		\subsubsection{Qualitative Comparisons} 
		The inpainting results on FFHQ~\cite{karras2019style}, Places2~\cite{zhou2017places}, and ImageNet~\cite{deng2009imagenet} of different methods are shown in \Fref{figure: inpainting_results_256}. Overall, PUT is more capable of catching the structure of the content in the image than other methods when most pixels are missing. It is clear that: 1) most existing methods perform better on monotonous datasets (\textit{e.g.}, FFHQ for faces and Places2 for natural scenes) than complex datasets (\textit{i.e.}, ImageNet), indicating their poor generalization ability. By contrast, our PUT generates vivid results for different cases and different datasets. 2) diverse and photorealistic results are produced by PUT. However, the results generated by other methods are full of artifacts, especially for the cases from ImageNet.

		\subsubsection{User Study} 
		For the evaluation of subjective quality, a user study is conducted. Since PUT is mainly designed for pluralistic image inpainting, only pluralistic methods are evaluated. Specifically, we randomly sample 20 pairs of images and masks from the test set of each dataset. For each pair, we generate one inpainted image using each method and ask the participants to rank these images according to their photorealism from high to low. We calculate the ratio of each method among the rank 1st images. Results are shown in \Fref{figure: usr_study}. Our method takes up at least 30\% of the rank 1st images and outperforms other baseline methods.

		\begin{figure*}
			\centering
			\includegraphics[width=2.05\columnwidth]{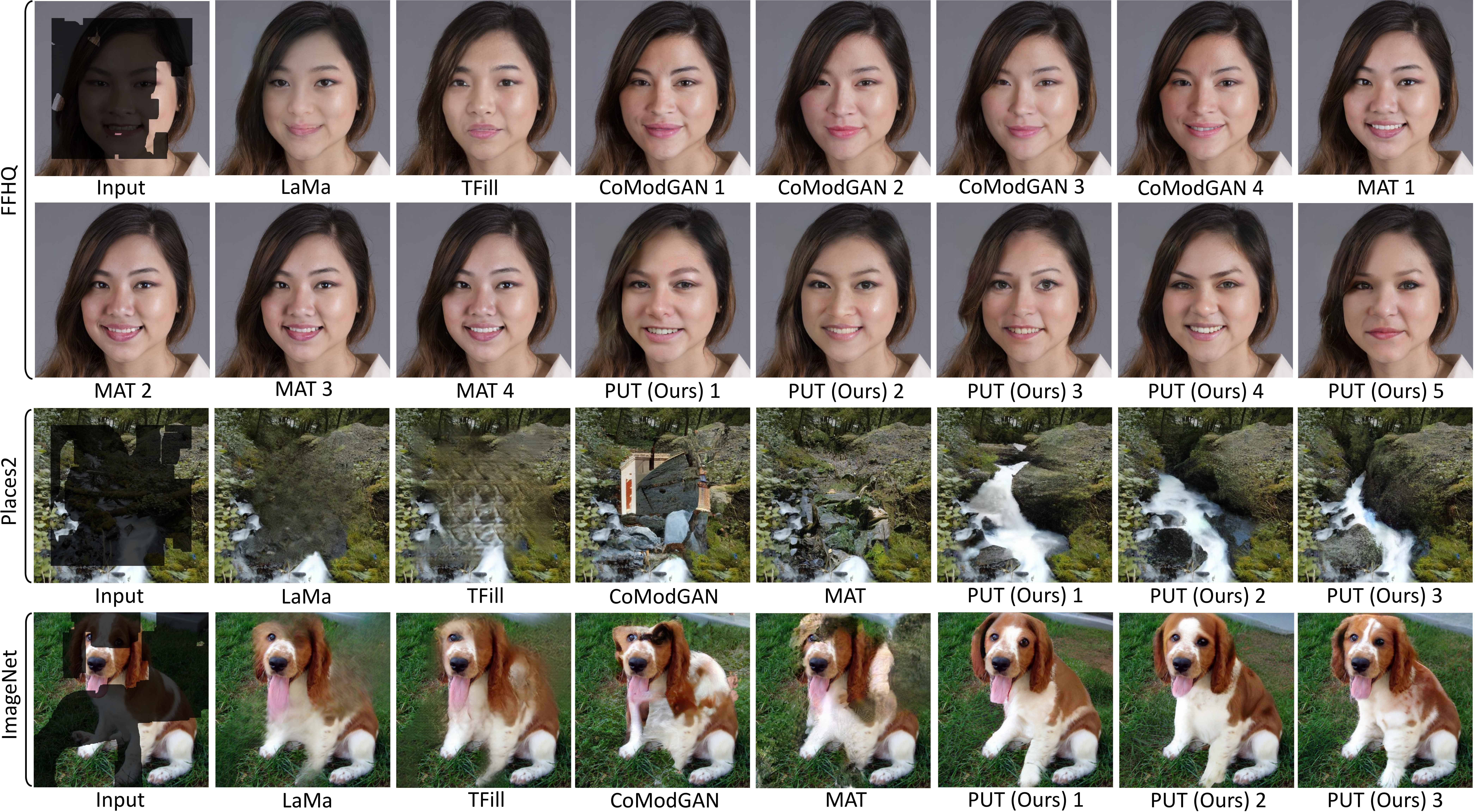} 
			\caption{Results produced by different methods. 
				PUT generates more photorealism images than LaMa~\cite{suvorov2022resolution} and TFill~\cite{zheng2022bridging}, and produces more diverse results than CoModGAN~\cite{zhao2021comodgan} and MAT~\cite{li2022mat}. For Places2 and ImageNet, only one sample of CoModGAN and MAT is presented due to their low diversity and limited space. Shown at $512 \times 512$ resolution.}
			\label{figure: inpainting_results_512}
		\end{figure*}

			\begin{figure}[t]
		\centering
		\scriptsize
		\setlength\tabcolsep{2pt}
	\begin{tabular}{cccc}
		\rotatebox{90}{FID$\downarrow$}  &
		\multicolumn{1}{m{0.305\linewidth}}{\includegraphics[width=1\linewidth]{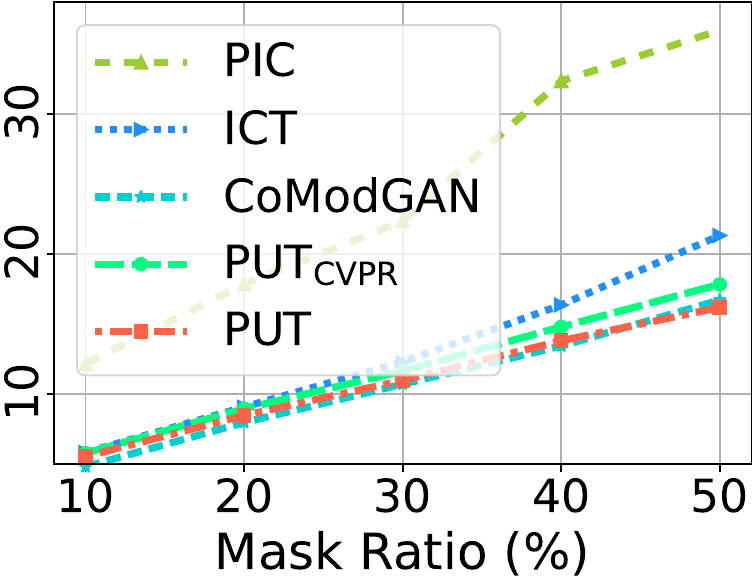}} &
		\multicolumn{1}{m{0.305\linewidth}}{\includegraphics[width=1\linewidth]{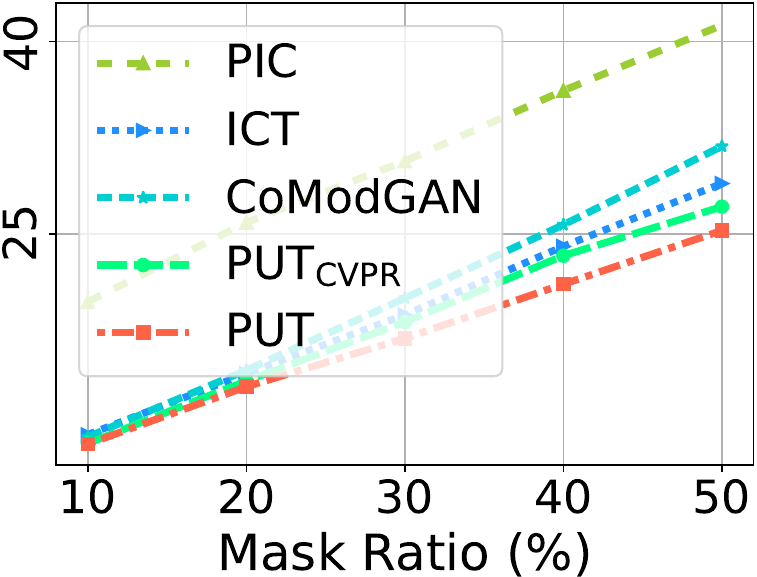}} &
		\multicolumn{1}{m{0.305\linewidth}}{\includegraphics[width=1\linewidth]{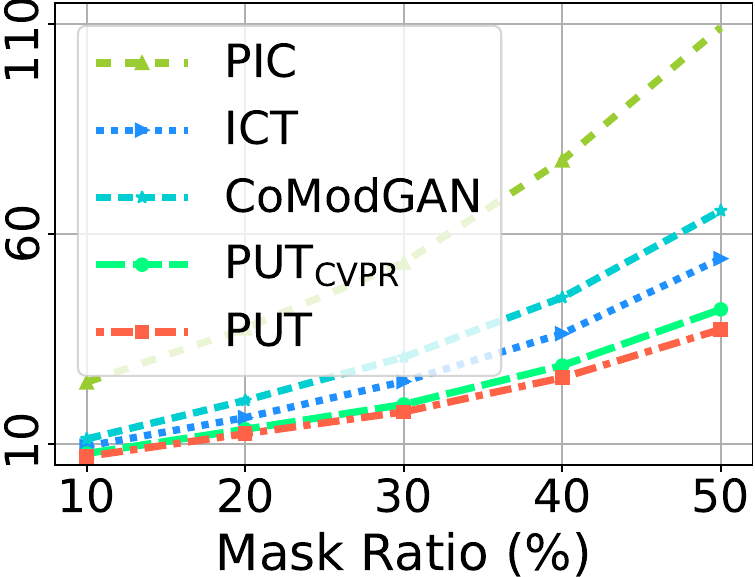}}\\
		
		\rotatebox{90}{LPIPS$\uparrow$}  &
		\multicolumn{1}{m{0.305\linewidth}}{\includegraphics[width=1\linewidth]{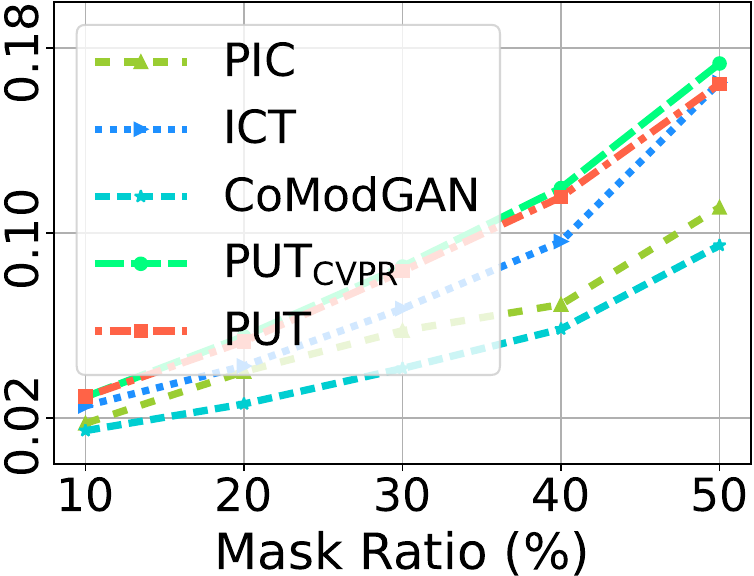}} &
		\multicolumn{1}{m{0.305\linewidth}}{\includegraphics[width=1\linewidth]{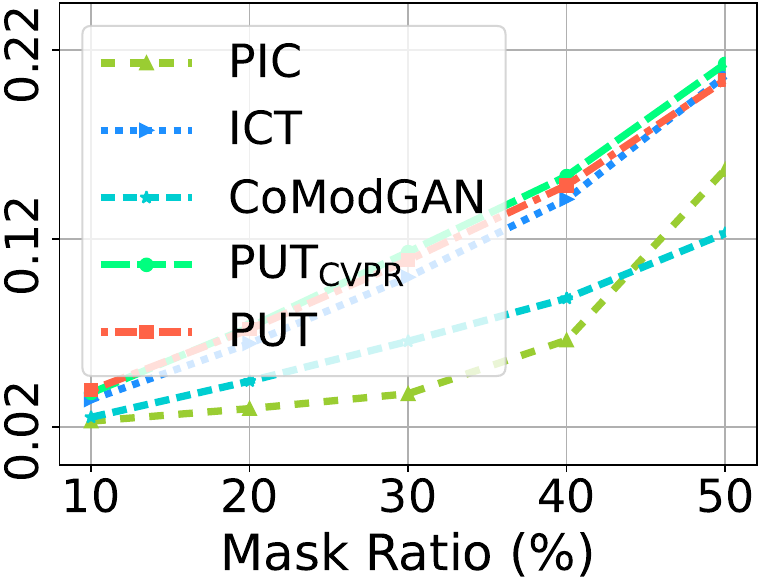}} &
		\multicolumn{1}{m{0.305\linewidth}}{\includegraphics[width=1\linewidth]{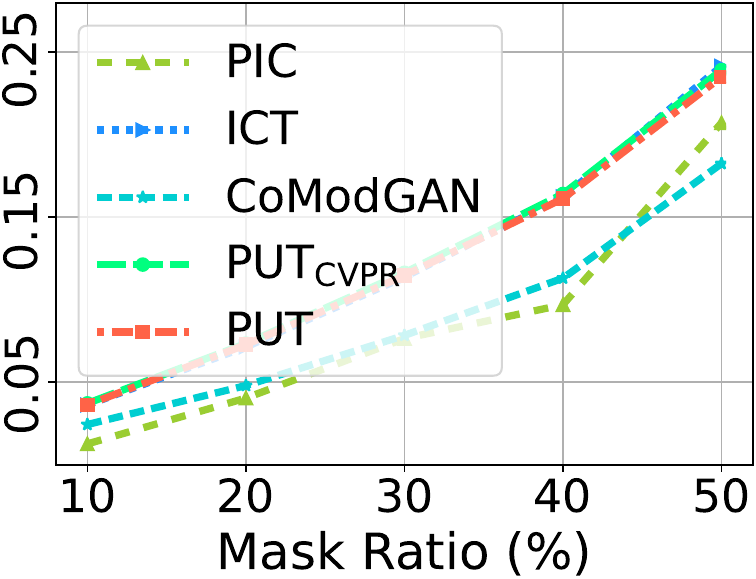}}\\
		
		& \multicolumn{1}{c}{FFHQ~\cite{karras2019style}} & \multicolumn{1}{c}{Places2~\cite{zhou2017places}} & \multicolumn{1}{c}{ImageNet~\cite{deng2009imagenet}} \\
	\end{tabular}	
	\caption{Diversity and fidelity comparison between different pluralistic methods. RePaint~\cite{lugmayr2022repaint} is not evaluated due to its slow inference speed (about 6 minutes per image). Reported on  $256 \times 256$ resolution.}
	\label{fig: lpips_and_fid_256}
	\vspace{-5pt}
	\end{figure}

	\begin{figure}[t]
		\centering
		\scriptsize
		\setlength\tabcolsep{2pt}
	\begin{tabular}{cccc}
		\rotatebox{90}{FID$\downarrow$}  &
		\multicolumn{1}{m{0.305\linewidth}}{\includegraphics[width=1\linewidth]{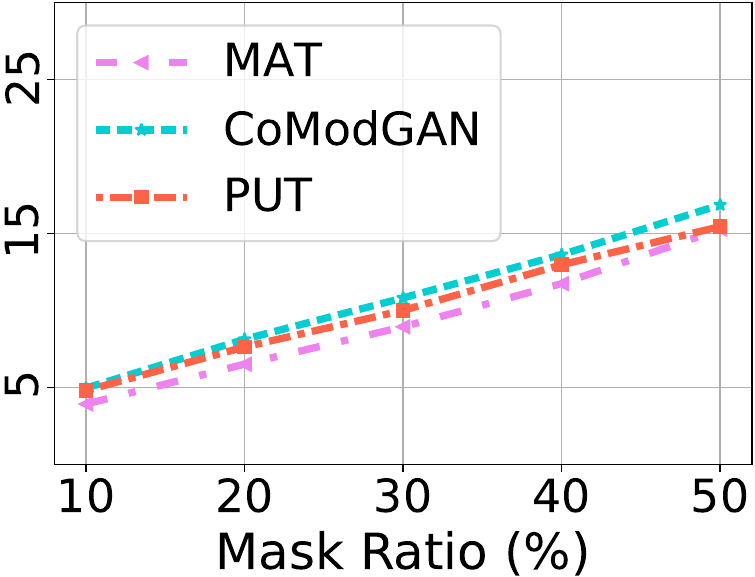}} &
		\multicolumn{1}{m{0.305\linewidth}}{\includegraphics[width=1\linewidth]{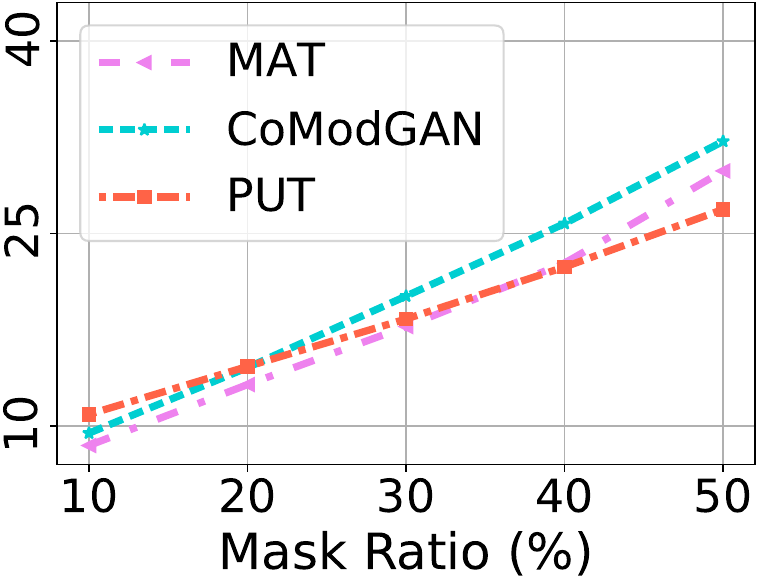}} &
		\multicolumn{1}{m{0.305\linewidth}}{\includegraphics[width=1\linewidth]{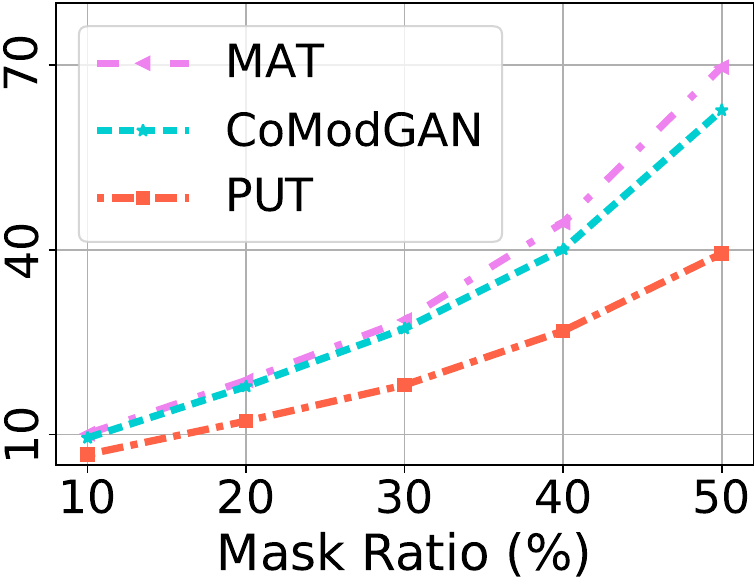}}\\
		
		\rotatebox{90}{LPIPS$\uparrow$}  &
		\multicolumn{1}{m{0.305\linewidth}}{\includegraphics[width=1\linewidth]{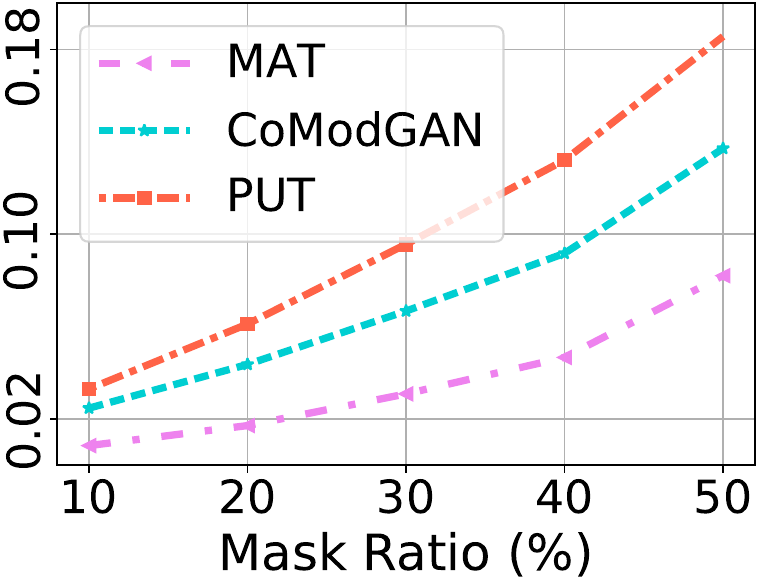}} &
		\multicolumn{1}{m{0.305\linewidth}}{\includegraphics[width=1\linewidth]{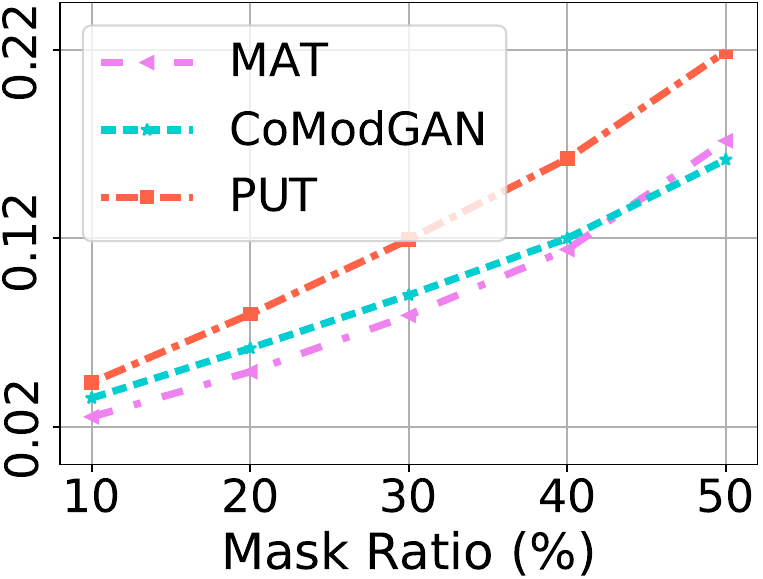}} &
		\multicolumn{1}{m{0.305\linewidth}}{\includegraphics[width=1\linewidth]{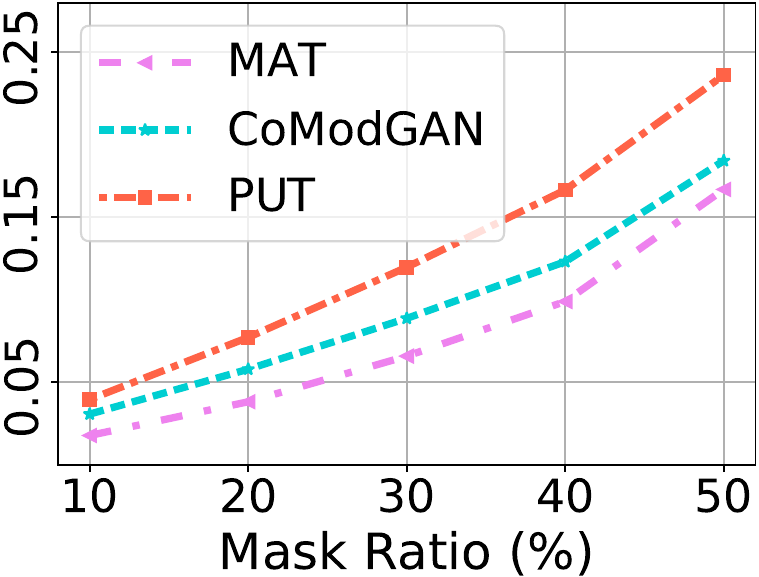}}\\

		& \multicolumn{1}{c}{FFHQ~\cite{karras2019style}} & \multicolumn{1}{c}{Places2~\cite{zhou2017places}} & \multicolumn{1}{c}{ImageNet~\cite{deng2009imagenet}} \\
	\end{tabular}	
	\caption{Diversity and fidelity comparison between different pluralistic methods. Reported on  $512 \times 512$ resolution.}
	\label{fig: lpips_and_fid_512}
	\vspace{-5pt}
	\end{figure}

		\subsubsection{Quantitative Comparisons}
		\label{sec: quantitative_comparison_256}
		We further demonstrate the superiority of PUT in terms of diversity and fidelity over other pluralistic methods. Specifically, the mean LPIPS distance~\cite{zhang2018unreasonable} between pairs of randomly generated results for the same input image is calculated. Following ICT~\cite{wan2021high}, five pairs per input image are generated. Meanwhile, the Fréchet Inception Distance (FID)~\cite{heusel2017gans} is also computed between the inpainted images and ground-truth images to reveal the fidelity of generated results. The curves of LPIPS and FID are shown in \Fref{fig: lpips_and_fid_256}. It can be seen that: 1) PUT achieves the best fidelity on Places2~\cite{zhou2017places} and ImageNet~\cite{deng2009imagenet}, especially for large areas of masked regions. For FFHQ~\cite{karras2019style}, CoModGAN~\cite{zhao2021comodgan} achieves overall better fidelity (lower FID) than PUT. This is in accordance with our expectation since CoModGAN is implemented based on StyleGAN2~\cite{karras2020analyzing}, which is a powerful face generator. 2) PUT achieves the best diversity (highest LPIPS) on all datasets. As ICT, PUT$_{\rm CVPR}$, and PUT are all transformer based autoregressive methods, they have similar capabilities of generating diverse results. However, PUT samples the tokens for multiple masked patches (20, more exactly) per iteration, while ICT and PUT$_{\rm CVPR}$ sample the token for one masked patch per iteration. This difference makes PUT 20× faster during inference (Ref. \Tref{table:flops_and_params}).

		\begin{table*}
			\setlength{\tabcolsep}{8pt}
			\centering
			\caption{Quantitative results for resolution $512 \times 512$. 
			PUT samples the tokens of all and 20 masked patches for comparison with deterministic and pluralistic methods, respectively.  \textcolor{Gray}{Gray} results indicate more training data is used.}
			\begin{tabular}{c|c|ccc|ccc|ccc}
				\hline
				\multicolumn{2}{c|}{Dataset} & \multicolumn{3}{c|}{FFHQ~\cite{karras2019style}} & \multicolumn{3}{c|}{Places2~\cite{zhou2017places}} &\multicolumn{3}{c}{ImageNet~\cite{deng2009imagenet}} \\ 
				\hline
				\multicolumn{2}{c|}{Mask Ratio (\%)} & 20-40 & 40-60 & 10-60 & 20-40 &40-60 & 10-60 & 20-40 &40-60 & 10-60\\
				\hline
				\multirow{6}*{\rotatebox{0}{FID$\downarrow$}} 
				& LaMa (WACV, 2022)~\cite{suvorov2022resolution} &\underline{\textbf{9.523}} &18.876 &\underline{\textbf{11.765}} &\underline{\textbf{\textcolor{Gray}{17.031}}} &\textcolor{Gray}{31.182} &\underline{\textbf{\textcolor{Gray}{20.663}}} &18.754 &44.579 &26.633\\
				&TFill (CVPR, 2022)~\cite{zheng2022bridging} &12.050 &21.186 &14.099 &24.619 &40.502 &28.448 &27.163 &58.695 &36.314  \\
				& PUT (Ours) &10.255 &\underline{\textbf{18.694}} &12.239 &20.156 &\underline{\textbf{31.159}} &22.710 &\underline{\textbf{18.619}} &\underline{\textbf{42.459}} &\underline{\textbf{25.452}} \\
				&\underline{\textbf{25.452}} \\
				\cline{2-11}
				& CoModGAN (ICLR, 2021)~\cite{zhao2021comodgan} &10.246 &17.933 &12.495 &\textcolor{Gray}{19.591} &\textcolor{Gray}{33.147} &\textcolor{Gray}{23.471} &25.582 &57.126 &34.761  \\
				& MAT (CVPR, 2022)~\cite{li2022mat} &\underline{\textbf{8.269}} &\underline{\textbf{15.578}} &\underline{\textbf{10.124}} &\underline{\textbf{\textcolor{Gray}{17.850}}} &\textcolor{Gray}{31.184} &\underline{\textbf{\textcolor{Gray}{21.655}}} &26.752 &63.159 &36.918\\
				& PUT (Ours) &10.806 &18.585 &12.471 &20.513 &\underline{\textbf{30.914}} &23.315 & \underline{\textbf{19.651}} &\underline{\textbf{41.373}} &\underline{\textbf{25.675}}  \\ 
				
				\hline

				\multirow{6}*{\rotatebox{0}{LPIPS$\downarrow$}} 
				& LaMa (WACV, 2022)~\cite{suvorov2022resolution}  &0.1047 &0.1964 &\underline{\textbf{0.1327}} &\textcolor{Gray}{0.1642} &\underline{\textbf{\textcolor{Gray}{0.2339}}} &\underline{\textbf{\textcolor{Gray}{0.1642}}} &\underline{\textbf{0.1228}} &\underline{\textbf{0.2332}} &\underline{\textbf{0.1583}}\\
				&TFill (CVPR, 2022)~\cite{zheng2022bridging} &\underline{\textbf{0.0942}} &0.2116 &0.1462 &\underline{\textbf{0.1576}} &0.2650 &0.1889 &0.1528 &0.2628 &0.1860\\
				& PUT (Ours) &0.1089 &\underline{\textbf{0.1944}} &0.1335 &0.1927 &0.2815 &0.2193 &0.1280 &0.2357 &0.1621 \\
				\cline{2-11}
				& CoModGAN (ICLR, 2021)~\cite{zhao2021comodgan} &0.1194 &0.2146 &0.1472 &\textcolor{Gray}{0.1462} &\textcolor{Gray}{0.2511} &\textcolor{Gray}{0.1772} &0.1523 &0.2709 &0.1887 \\
				& MAT (CVPR, 2022)~\cite{li2022mat} &\underline{\textbf{0.0942}} &\underline{\textbf{0.1797}} &\underline{\textbf{0.1200}} &\underline{\textbf{\textcolor{Gray}{0.1351}}} &\underline{\textbf{\textcolor{Gray}{0.2393}}} &\underline{\textbf{\textcolor{Gray}{0.1668}}} &0.1552 &0.2790 &0.1928 \\
				& PUT (Ours) &0.1147 &0.2039 &0.1403 &0.1969 &0.2876 &0.2303 &\underline{\textbf{0.1333}} &\underline{\textbf{0.2429}} &\underline{\textbf{0.1677}}  \\
				\hline

				\multirow{6}*{\rotatebox{0}{PSNR$\uparrow$}} 
				& LaMa (WACV, 2022)~\cite{suvorov2022resolution}  &\underline{\textbf{29.006}} &24.012 &\underline{\textbf{27.863}} &\textcolor{Gray}{26.249} &\textcolor{Gray}{22.691} &\underline{\textbf{\textcolor{Gray}{25.469}}} &\underline{\textbf{26.394}} &\underline{\textbf{21.840}} &\underline{\textbf{25.287}} \\
				&TFill (CVPR, 2022)~\cite{zheng2022bridging} &27.807 &23.285 &26.815 &\underline{\textbf{26.699}} &\underline{\textbf{22.741}} &25.449 &25.206 &21.134 &24.250 \\
				& PUT (Ours) &28.699 &\underline{\textbf{24.227}} &27.755 &24.009 &21.051 &23.346 &25.718 &21.338 &24.709 \\
				\cline{2-11}
				& CoModGAN (ICLR, 2021)~\cite{zhao2021comodgan} &27.325 &22.439 &26.220 &\textcolor{Gray}{24.776} &\underline{\textbf{\textcolor{Gray}{20.969}}} &\textcolor{Gray}{23.886} &24.223 &19.659 &23.114\\
				& MAT (CVPR, 2022)~\cite{li2022mat} &\underline{\textbf{28.270}} &\underline{\textbf{23.375}} &\underline{\textbf{27.218}} &\underline{\textbf{\textcolor{Gray}{24.785}}} &\textcolor{Gray}{20.870} &\underline{\textbf{\textcolor{Gray}{23.899}}} &24.249 &19.798 &23.255 \\
				& PUT (Ours) &27.700 &23.109 &26.724 &23.240 &20.555 &22.800 &\underline{\textbf{24.814}} &\underline{\textbf{20.405}} &\underline{\textbf{23.763}} \\
				\hline
				
				\multirow{6}*{\rotatebox{0}{SSIM$\uparrow$}} 
				& LaMa (WACV, 2022)~\cite{suvorov2022resolution}  &0.937 &0.837 &0.901 &\textcolor{Gray}{0.869} &\underline{\textbf{\textcolor{Gray}{0.727}}} &\underline{\textbf{\textcolor{Gray}{0.822}}}  &\underline{\textbf{0.898}} &\underline{\textbf{0.752}} &\underline{\textbf{0.845}}\\
				&TFill (CVPR, 2022)~\cite{zheng2022bridging} &\underline{\textbf{0.923}} &\underline{\textbf{0.817}} &\underline{\textbf{0.887}} &\underline{\textbf{0.887}} &0.722 &0.817 &0.876 &0.726 &0.824 \\
				& PUT (Ours) &0.936 &0.853 &0.908 &0.769 &0.637 &0.723 &0.890 &0.750 &0.841 \\
				\cline{2-11}
				& CoModGAN (ICLR, 2021)~\cite{zhao2021comodgan} &\underline{\textbf{0.918}} &\underline{\textbf{0.801}} &\underline{\textbf{0.878}} &\textcolor{Gray}{0.839} &\underline{\textbf{\textcolor{Gray}{0.676}}} &\underline{\textbf{\textcolor{Gray}{0.785}}} &0.861 &0.688 &0.800 \\
				& MAT (CVPR, 2022)~\cite{li2022mat} &0.932 &0.833 &0.897 &\underline{\textbf{\textcolor{Gray}{0.840}}} &\underline{\textbf{\textcolor{Gray}{0.676}}} &\underline{\textbf{\textcolor{Gray}{0.785}}} &0.862 &0.689 &0.820 \\
				& PUT (Ours) &0.922 &0.820 &0.888 &0.759 &0.621 &0.700 &\underline{\textbf{0.874}} &\underline{\textbf{0.718}} &\underline{\textbf{0.819}}\\
				\hline
			\end{tabular}
			\label{tab: results_with_resolution_512}
		\end{table*}

		\begin{figure*}[t]
			\centering
			\includegraphics[width=2.0\columnwidth]{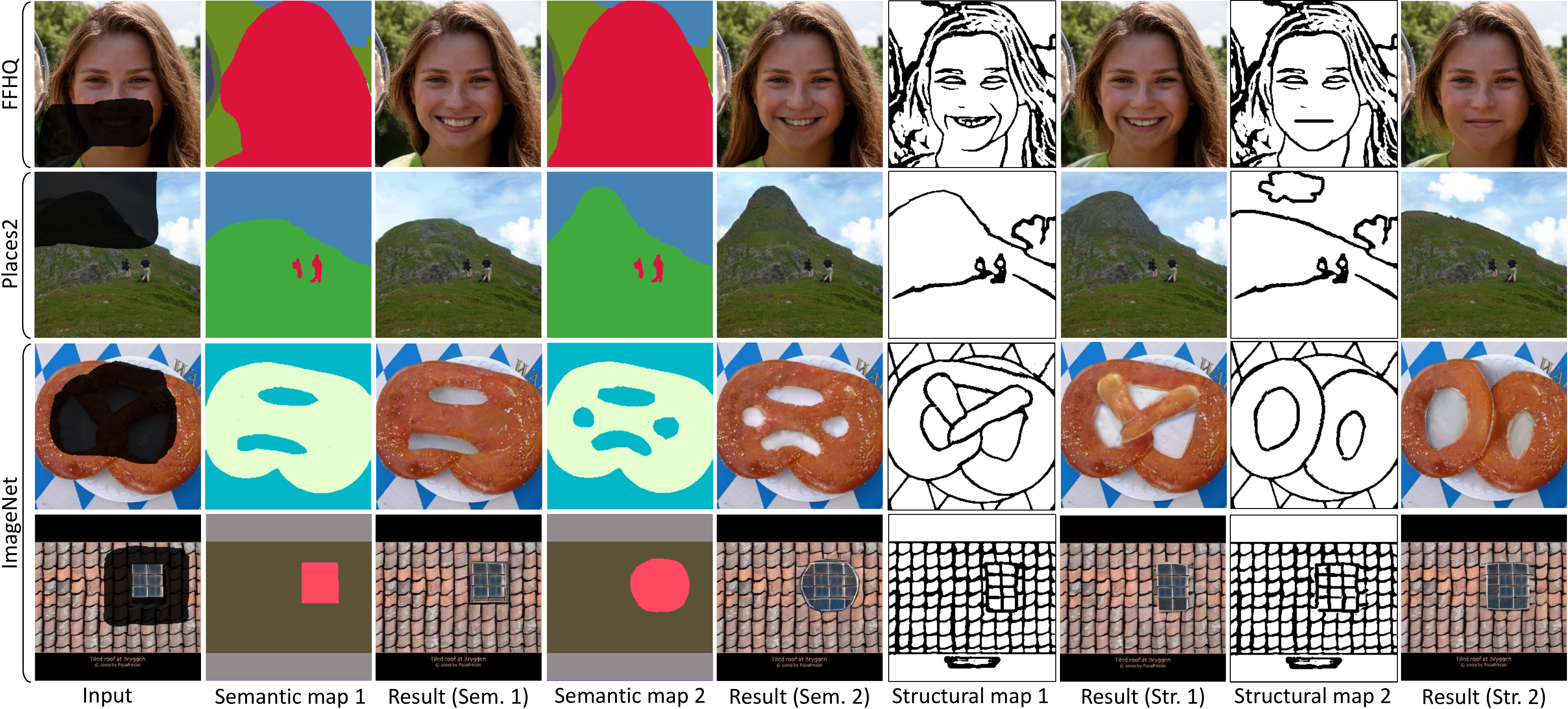} 
			\mycaption{Inpainted results with different conditions. PUT strictly follows different types of conditions. Shown at $256 \times 256$.}
			
			\label{figure: controllable_inpainting_results}
		\end{figure*}

	In \Tref{table:quantitative_results_256}, we compare different methods in terms of several metrics. Only one recovered output is produced for each input. For the comparison with pluralistic methods, PUT performs the best in all metrics on ImageNet and achieves the best scores in almost all metrics on Places2. Specifically, the FID score of PUT on ImageNet with mask ratio 40\%-60\% is 15.269 lower than ICT and 4.824 lower than PUT$_{\rm CVPR}$. Though CoModGAN achieves a better FID score on FFHQ than PUT, the FID score of CoModGAN on ImageNet is much poorer than PUT. This may be due to the inherent properties in the model structure of CoModGAN, which is similar to the powerful face generator StyleGAN2~\cite{karras2020analyzing}. For the comparisons with deterministic methods, PUT also achieves almost the best FID score on Places2 and ImageNet.

		\subsection{Evaluation on $\mathbf{512\times 512}$ Resolution}
		\label{sec: comparision_512}
		As some recently proposed methods support image inpainting on $512 \times 512$ resolution, we also apply PUT to such resolution. The evaluated methods are LaMa~\cite{suvorov2022resolution}, TFill~\cite{zheng2022bridging}, CoModGAN~\cite{zhao2021comodgan}, and MAT~\cite{li2022mat}. \qiankun{Note that the models of LaMa, CoModGAN, and MAT on ImageNet are trained by ourselves since there are no available models.}

		\subsubsection{Qualitative Comparisons}
		The inpainting results are shown in \Fref{figure: inpainting_results_512}. Overall, PUT produces diverse and promising results on different datasets. Benefiting from StyleGAN2, both CoModGAN and MAT produce impressive face images. However, their diversity is very limited. In addition, they fail to get plausible results on Places2 and ImageNet.

		\begin{figure*}[t]
			\centering
			\includegraphics[width=2.00\columnwidth]{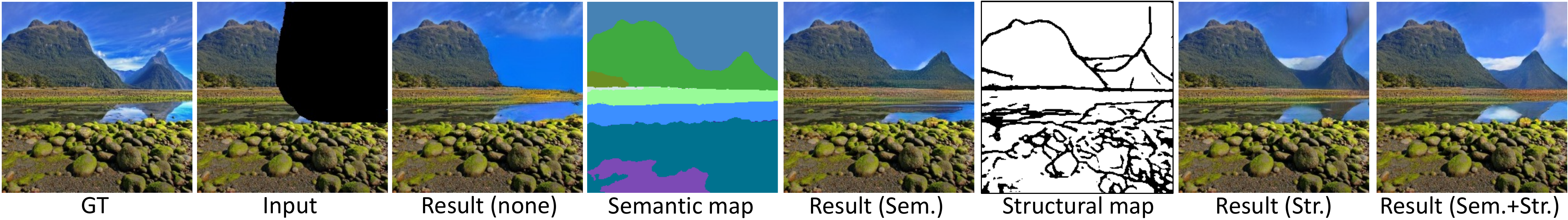} 
			\caption{\qiankun{The user-provided conditions can avoid some artifacts and make the results more desirable. Shown at $256 \times 256$.}}
			\label{figure: controllability_avoid_artifacts}
		\end{figure*}

		\subsubsection{Quantitative Comparisons}
	Similar to the settings in \Sref{sec: quantitative_comparison_256}, we compare different pluralistic methods in terms of fidelity and diversity in \Fref{fig: lpips_and_fid_512}. Compared to MAT, PUT achieves comparable fidelity on FFHQ and Places2 but much higher diversity.

	In \Tref{tab: results_with_resolution_512}, we present several metrics of different methods. Compared to deterministic methods, PUT achieves the best FID scores on different datasets when 40\%-60\% of pixels are masked. Compared to pluralistic methods, both CoModGAN and MAT perform overall better than PUT on FFHQ. We argue that the superiority of these two methods comes from the well-designed model structure in StyleGAN2~\cite{karras2020analyzing} for face generation. On the more complex dataset (\textit{i.e.}, ImageNet), PUT performs much better than them on all metrics. In addition, PUT achieves comparable performances to CoModGAN and MAT on Places2, even though the number of training images of PUT (0.238M) is smaller than that of CoModGAN (8M) and MAT (1.8M).

		\begin{table}
			\setlength{\tabcolsep}{0.1pt}
			\centering
			\mycaption{Quantitative results of PUT with different conditions. Images are sampled within one iteration.  Evaluated on $256\times256$ with mask ratio 10\%-60\%.}
			\mycolor
			\begin{tabular}{c|c|c|c|c|c}
				\hline
				 & \makecell{Trained with \\conditions} & \makecell{Conditions} & \multicolumn{1}{c|}{FFHQ\cite{karras2019style}} & \multicolumn{1}{c|}{Places2\cite{zhou2017places}} &\multicolumn{1}{c}{ImageNet\cite{deng2009imagenet}} \\ 
				
				\hline
				\multirow{5}*{\rotatebox{0}{FID$\downarrow$}} 
				& $\times$ & none & 12.728 & 21.234  &24.794 \\
				\cline{2-6}
				& \multirow{4}*{\rotatebox{0}{$\checkmark$}}
				& none & 12.778 & 21.851 & 26.402\\
				& & Sem. & 12.340 & 20.403 & 23.729 \\
				& & Str. & 11.257 & 19.250 & 18.281\\
				& & Sem.$+$Str. & \underline{\bf 11.117} & \underline{\bf 18.781} & \underline{\bf 17.916}\\
				\hline
				\multirow{5}*{\rotatebox{0}{mIoU$\uparrow$}} 
				& $\times$ & none & 0.7179 & 0.6911 & 0.5909 \\
				\cline{2-6}
				& \multirow{4}*{\rotatebox{0}{$\checkmark$}}
				& none & 0.7184 & 0.6786 & 0.5796\\
				& & Sem. & 0.7286 & \underline{\bf 0.7133} & 0.6153\\
				& & Str. & 0.7279 & 0.7083 & 0.6146\\
				& & Sem.$+$Str. & \underline{\bf 0.7338} & 0.7131 & \underline{\bf 0.6201}\\
				\hline
				\multirow{5}*{\rotatebox{0}{F$_{1}\uparrow$}} 
				& $\times$ & none & 0.7796 & 0.7197 & 0.7682 \\
				\cline{2-6}
				& \multirow{4}*{\rotatebox{0}{$\checkmark$}}
				& none & 0.7785& 0.7119 & 0.7638\\
				& & Sem. & 0.7881 & 0.7376 & 0.7793\\
				& & Str. & 0.8237 & 0.7751 & 0.8207\\
				& & Sem.$+$Str. & \underline{\bf 0.8240} & \underline{\bf 0.7763} & \underline{\bf 0.8215}\\
				\hline
			\end{tabular}
			\label{tab: controllable_inpainting_quantitative_results_256}
		\end{table}

		\qiankun{\subsection{Evaluation of Controllable Image Inpainting}}
		\label{sec: controllable inpainting}
		\qiankun{In this section, we show the controllability of PUT with the user-provided semantic and structural maps. Without loss of generality, the experiments in this section are conducted on $256 \times 256$ resolution.}
				
		\qiankun{\subsubsection{Qualitative Analysis}}
		\qiankun{The visual results are shown in \Fref{figure: controllable_inpainting_results}. As we can see, PUT follows the conditions provided by users, proving the superior controllability of PUT. For example, in the third scene, the donut is modified to have different shapes according to different semantic and structural maps. In \Fref{figure: controllability_avoid_artifacts}, we show a case where the artifacts are avoided with more conditions. It can be seen that the result is over-smoothed when no condition is provided (Result (none)). With the help of the semantic map, the visual quality is greatly improved (Result (Sem.)). However, the mountain's reflection in the water is unnatural. When PUT is conditioned on the structural map, the inpainted image (Result (str.)) follows the structural map but the generated content is still not perfect. Finally, when both semantic and structural guidances are given, the inpainting results are further improved (Result (Sem.+Str.)).}

\begin{figure}
		\centering
			\includegraphics[width=1.00\columnwidth]{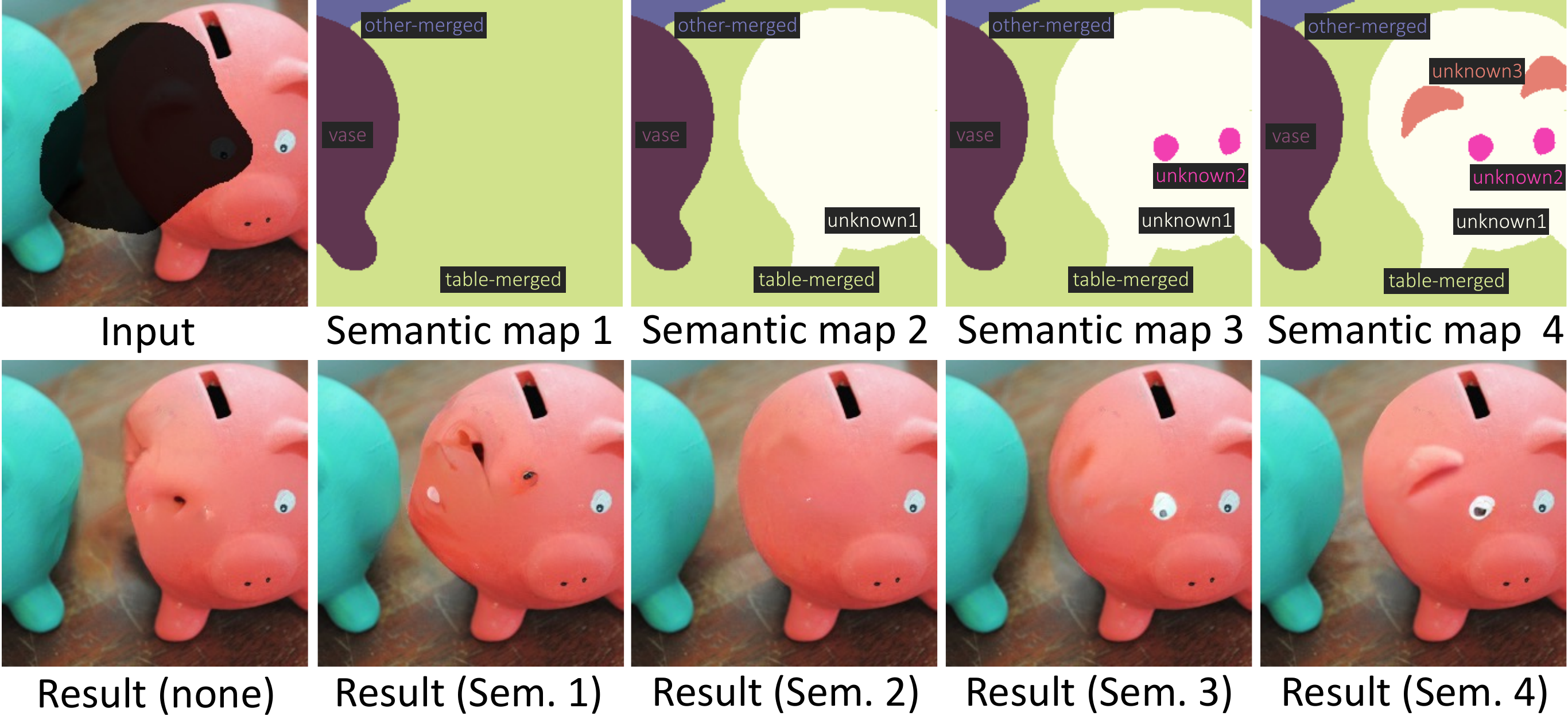} 
			\mycaption{The effectiveness of unknown category strategy. Semantic map 1 is obtained by Mask2Former~\cite{cheng2021mask2former}. Semantic maps 2-4 are manually added with 1-3 unknown categories.  Shown at $256 \times 256$.}
		\label{figure: unknown_category}
		\end{figure}

		\qiankun{The effectiveness of the unknown category strategy is shown in \Fref{figure: unknown_category}. It can be seen that the segmentation model trained with a pre-defined closed-set of categories cannot recognize some objects correctly. Conditioned on an improper semantic map (Semantic map 1), PUT generates an artificial inpainted image (Result (Sem. 1)). However, when unrecognized objects (\textit{e.g.}, piggy bank, eyes, ears) are labeled with unknown categories, images with better quality are produced.}

		\qiankun{\subsubsection{Quantitative Analysis}}
	\qiankun{As manually drawing the semantic and structural maps for all images in test splits is laborious and time-consuming, we use the maps obtained from the completed images to simulate user-provided maps for quantitative evaluation. To evaluate how well PUT follows the provided conditions, we get the mIoU (mean Intersection over Union)/F1 score between the provided segmentation/sketch maps and the segmentation/sketch maps obtained from inpainted images.}

		\begin{table*}
			\setlength{\tabcolsep}{1.5pt}
			\centering
			\caption{The comparison of FLOPs, number of parameters and inference time. 
				For ICT~\cite{wan2021high}, PUT$_{\rm CVPR}$~\cite{liu2022reduce} and PUT, these metrics of transformer and other components are shown separately since only the transformer need to be iteratively used for pluralistic image inpainting.
				Due to the various dependencies of different codes, it is hard to make a fair comparison of inference time. The used GPUs are provided for reference. Note that the Mask Ratio (MR) lies in the open interval (0,1).}  
			\label{table:flops_and_params}
			
			\begin{tabular}{c|cc|c|c|c|c|c|c}
				\hline 
				Resolution &\multicolumn{2}{c|}{Method}&Pluralistic  &\makecell{Parameters \\ (M)} &\makecell{FLOPs \\ (G)}& \makecell{Time \\(s/image)}  & \makecell{Number of \\Iterations }& GPU \\
				\hline
				\multirow{17}*{$256\times256$} 
				& \multicolumn{2}{c|}{EdgeConnect (ICCVW, 2019)~\cite{nazeri2019edgeconnect}} &$\times$ &20.5 &122.6  &0.024 & 1 &TITAN Xp\\
				& \multicolumn{2}{c|}{MEDFE (ECCV, 2020)~\cite{liu2020rethinking}} &$\times$ &124.3 & 138.0  &0.127 & 1 &TITAN Xp\\
				& \multicolumn{2}{c|}{LaMa (WACV, 2022)~\cite{suvorov2022resolution}}&$\times$ &22.9 &42.9 &0.029 & 1 &RTX 3090\\
				&\multicolumn{2}{c|}{TFill (CVPR, 2022)~\cite{zheng2022bridging}}&$\times$ &61.2 &28.3 &0.046   &1 &RTX 3090\\ 
				& \multicolumn{2}{c|}{PIC (CVPR, 2019)~\cite{zheng2019pluralistic}} &$\checkmark$ &5.7  &36.2 &0.035 & 1 &TITAN Xp\\
				
				& \multicolumn{2}{c|}{CoModGAN (ICLR, 2021)~\cite{zhao2021comodgan}}  &$\checkmark$ &75.5 &14.8 &0.038 & 1&TITAN Xp\ \\
				&\multicolumn{2}{c|}{RePaint (CVPR, 2022)~\cite{lugmayr2022repaint}} &$\checkmark$ &527.2 &1113.8$\times$Iters. &0.079$\times$Iters.  &4570 &RTX 3090\\
				\cdashline{2-9}[3pt/1.5pt]
				& ICT (ICCV, 2021)~\cite{wan2021high} &\makecell{\textcolor{gray}{FFHQ~\cite{karras2019style}} \\\textcolor{gray}{Places2~\cite{zhou2017places}}\\\textcolor{gray}{ImageNet~\cite{deng2009imagenet}}}&$\checkmark$ & \makecell{91.8+10.3 \\ 106.2+10.3 \\422.4+10.3}  & \makecell{957.5$\times$Iters.+62.0 \\ 283.6$\times$Iters.+62.0 \\791.5$\times$Iters.+62.0} &\makecell{0.073$\times$Iters.+0.010\\0.031$\times$Iters.+0.010\\0.058$\times$Iters.+0.010}  & \makecell{ $\lceil{\rm MR} \times {\rm 4096} \rceil$ \\$\lceil{\rm MR} \times {\rm 1024} \rceil$ \\$\lceil{\rm MR} \times {\rm 1024} \rceil$} &RTX 3090\\
				\cdashline{2-9}[3pt/1.5pt]
				& PUT$_{\rm CVPR}$ (CVPR, 2022)~\cite{liu2022reduce} &\makecell{\textcolor{gray}{FFHQ~\cite{karras2019style}} \\\textcolor{gray}{Places2~\cite{zhou2017places}}\\\textcolor{gray}{ImageNet~\cite{deng2009imagenet}}}&$\checkmark$ &  \makecell{90.5+7.6 \\ 105.6+7.6 \\421.2+7.6}  & \makecell{217.2$\times$Iters.+31.4 \\ 257.7$\times$Iters.+31.4 \\765.6$\times$Iters.+31.4} &\makecell{0.026$\times$Iters.+0.015\\0.031$\times$Iters.+0.015\\0.058$\times$Iters.+0.015} &$\lceil{\rm MR} \times {\rm 1024} \rceil$ &RTX 3090\\
				\cdashline{2-9}[3pt/1.5pt]
				& \multicolumn{2}{c|}{PUT, no condition, (Ours) }&$\checkmark$ & 87.3+10.6  & 193.2$\times$Iters.+40.1 &0.026$\times$Iters.+0.016 &$\lceil {\rm MR}\times\frac{1024}{20}\rceil$ &\multirow{3}*{RTX 3090} \\
				& \multicolumn{2}{c|}{\qiankun{PUT, one condition, (Ours)}} &\qiankun{$\checkmark$} & \qiankun{87.3+11.2}  & \qiankun{193.2$\times$Iters.+40.6} &\qiankun{0.026$\times$Iters.+0.018} &\qiankun{$\lceil {\rm MR}\times\frac{1024}{20}\rceil$} & \\
				& \multicolumn{2}{c|}{\qiankun{PUT, two conditions, (Ours)} } &\qiankun{$\checkmark$} & \qiankun{87.3+11.8}  & \qiankun{193.2$\times$Iters.+41.1} &\qiankun{0.026$\times$Iters.+0.020} &\qiankun{$\lceil {\rm MR}\times\frac{1024}{20}\rceil$} &\\
				\hline

				\multirow{5}*{\rotatebox{0}{$512 \times 512$}} 
				& \multicolumn{2}{c|}{LaMa (WACV, 2022)~\cite{suvorov2022resolution}} &$\times$ &25.8 & 171.3 &0.033  &1 &RTX 3090\\
				&\multicolumn{2}{c|}{TFill (CVPR, 2022)~\cite{zheng2022bridging}} &$\times$ &104.4 & 50.5 &0.074 &1 &RTX 3090\\
				& \multicolumn{2}{c|}{CoModGAN (ICLR, 2021)~\cite{zhao2021comodgan}}  &$\checkmark$ &76.1 & 20.2 &0.061 &1 &TITAN Xp\\
				& \multicolumn{2}{c|}{MAT(CVPR, 2022)~\cite{li2022mat}} &\checkmark &58.7 &212.0 &0.065 &1 &RTX 3090\\
				& \multicolumn{2}{c|}{PUT (Ours)} &$\checkmark$ &87.3+20.5 &193.2$\times$Iters.+184.2 &0.026$\times$Iters.+ 0.018  &$\lceil {\rm MR}\times\frac{1024}{20}\rceil$ &RTX 3090 \\ 
				
				\hline 
			\end{tabular}
			\label{tab: flops_parameters_time}
		\end{table*}

		\begin{table}
			\setlength{\tabcolsep}{4pt}
			\centering
			\mycolor
			\mycaption{The effectiveness of unknown category strategy. Evaluated on $256\times256$ with mask ratio 10\%-60\%. }
			\begin{tabular}{c|cc|cc|cc}
				\hline
				\multirow{2}*{Configuration} &\multicolumn{2}{c|}{FFHQ\cite{karras2019style}} & \multicolumn{2}{c|}{Places2\cite{zhou2017places}} &\multicolumn{2}{c}{ImageNet\cite{deng2009imagenet}} \\ 
				\cline{2-7}
				& FID$\downarrow$ &PSNR$\uparrow$ & FID$\downarrow$ &PSNR$\uparrow$ &FID$\downarrow$ &PSNR$\uparrow$\\
				\hline
				Sem.$^{133}$ & \underline{\bf{12.340}} & \underline{\bf{27.980}} & 20.403 & \underline{\bf{26.066}} & 23.729 & \underline{\bf{25.211}} \\
				Sem.$^{113}$ & 13.074 & 27.400 & 20.927 & 25.999 & 24.877 & 24.983 \\
				Sem.$^{113}+$Unk. & 12.343 & 27.977 & \underline{\bf{20.387}} & \underline{\bf{26.066}} & \underline{\bf{23.674}} & 25.207\\
				\hline
			\end{tabular}
			\label{tab: ablate_unknown_strategy_256}
		\end{table}

		\qiankun{We first show the controllability of PUT with different conditions in \Tref{tab: controllable_inpainting_quantitative_results_256}. When no conditions are provided in the inference stage, comparable performances are achieved by two variants of PUT, \textit{i.e.}, trained with and without conditions, demonstrating that the placeholder embeddings for different conditions have no negative effect on the inpainting quality. For the PUT trained with conditions, all metrics are improved even when only one type of condition is provided, demonstrating the effectiveness of our controllable design and the effectiveness of placeholder embeddings in handling the absence of conditions. Interestingly, the structural maps result in a better FID score than semantic maps since they contain more detailed textures. Compared to only giving one type of condition, PUT achieves better metrics when two types of conditions are both utilized.}

		\qiankun{To qualitatively show the effectiveness of the unknown category strategy, we compare the models trained with different settings, including using all the categories recognized by Mask2Former~\cite{cheng2021mask2former} (denoted as Sem.$^{133}$), removing 20 categories (denotes as Sem.$^{113}$) recognized by Mask2Former, and using 20 unknown categories to substitute the removed categories (denoted as Sem.$^{113}+$Unk.). Only the FID and PSNR metrics between inpainted and ground-truth images are evaluated. The mIoU is not used since the sets of categories in these three settings are different. Results are shown in \Tref{tab: ablate_unknown_strategy_256}. Compared with Sem.$^{133}$, Sem.$^{113}$ achieves inferior performance on different metrics. When assisted with unknown categories, those unrecognized categories can be labeled with unknown categories, helping PUT inpaint images with better quality (Sem.$^{113}$ vs Sem.$^{113}+$Unk.).}

		\begin{table*}[h]
			\setlength{\tabcolsep}{5pt}
			\centering
			\caption{Quantitative results of different settings. For PVQVAE, the numbers of latent vectors in $\mathbf{e}$ and $\mathbf{e}'$ both are set to 512 and the Gumbel-softmax relaxation is not used in the training stage. For UQ-Transformer, the mask embedding is not introduced.	Images are sampled with $\mathcal{K}_1=1$ and $K_2=50$. Evaluated on $256\times256$ resolution with mask ratio 10\%-60\%.}  
			\begin{tabular}{c|l|cccc|cccc}
				\hline
				\multirow{2}*{Identifier} &\multicolumn{1}{c|}{\multirow{2}*{Configuration}}  & \multicolumn{4}{c|}{FFHQ~\cite{karras2019style}} & \multicolumn{4}{c}{Places2~\cite{zhou2017places}} \\ 
				\cline{3-10}
				& &FID$\downarrow$ &LPIPS$\downarrow$ &PSNR$\uparrow$ &SSIM$\uparrow$ &FID$\downarrow$ &LPIPS$\downarrow$ & PSNR$\uparrow$ &SSIM$\uparrow$\\
				\hline
				A & PUT in our conference paper~\cite{liu2022reduce} &14.554 &0.1231 &\underline{\bf 25.943}  &\underline{\bf 0.914} &\underline{\bf 22.121} &\underline{\bf 0.1569} &\underline{\bf 24.492} &\underline{\bf 0.806}\\ 
				B & A w/o Patched-based encoder &173.351 &0.3822 &12.360 &0.445 &179.294 &0.3895 &11.799 &0.373  \\
				C & A w/o Dual-codebook &\underline{\bf 13.960}&\underline{\bf 0.1201} &25.903 &0.906  &28.634 &0.1808 &23.600 &0.776 \\
				D & A w/o Guidance from reference branch in decoder &16.469 &0.1378 &25.547 &0.906 &25.084 &0.1773 &24.185 &0.798\\
				E & A w/o Un-quantized inputs to UQ-Transformer &26.098 &0.1654 &22.879 &0.843 &75.625 &0.2383 &19.429 &0.652\\
				F & A w/o Random quantization during training  &54.879 &0.2055 &23.174 &0.820 &44.588 &0.2075 &23.353 &0.763 \\
				\hline 
			\end{tabular}
			\label{tab: ablation_study_on_put_in_conference}
		\end{table*}

		\subsection{FLOPs, Parameters and Inference Time}
		From the results presented in  \Tref{table:flops_and_params}, we can see that for different methods, the FLOPs and the number of parameters are not constrained to be the same. For pluralistic methods, diffusion model based (\textit{e.g.}, RePaint) and transformer based autoregressive (like ICT, PUT$_{\rm CVPR}$ and PUT) solutions need several iterations to get diverse results, resulting in larger FLOPs and longer inference time. Compared with ICT and PUT$_{\rm CVPR}$, PUT has the smallest number of parameters and takes the least number of iterations, resulting in a much faster inference speed and much fewer FLOPs. For example, the number of iterations of PUT is about 5.0\% of ICT and PUT$_{\rm CVPR}$ and is at most 1.1\% of RePaint.

	\begin{table*}[t]
					\setlength{\tabcolsep}{0.5pt}
					\centering
					\caption{Quantitative results of different methods. We set $\mathcal{K}_1$ = 1 and increase $\mathcal{K}_2$ from 50 to 200 when the number of latent vectors in $\mathbf{e}$ is increased from 512 to 8192.  Except for the mentioned difference between different configurations, others remain the same. Evaluated at $256 \times 256$ resolution with mask ratio 10\%-60\%.}  
					\begin{tabular}{c|ccc|ccccc|ccccc}
							\hline
							\multirow{3}{*}{Identifier}&\multicolumn{3}{c|}{Configuration} & \multicolumn{5}{c|}{FFHQ~\cite{karras2019style}} & \multicolumn{5}{c}{Places2~\cite{zhou2017places}} \\ 
							\cline{2-14}
							&\makecell[c]{Number of\\latent vectors $K$}  &\makecell[c]{Gumbel-softmax\\ relaxation}
							&\makecell[c]{Mask\\Embedding} &\makecell[c]{Parameters of\\ UQ-Transformer} &FID$\downarrow$ &LPIPS$\downarrow$ &PSNR$\uparrow$ &SSIM$\uparrow$ &\makecell[c]{Parameters of\\ UQ-Transformer} &FID$\downarrow$ &LPIPS$\downarrow$ & PSNR$\uparrow$ &SSIM$\uparrow$\\
							\hline
							A &512  &$\times$ &$\times$ &90.5M &14.554 &0.1231 &25.943  &\underline{\bf 0.914} &105.6M &22.121 &0.1569 &24.492 &0.806\\ 
							G &512  &$\times$ &$\times$ &87.3M &14.764 &0.1238 &25.916 &0.905 &87.3M &22.696 &0.1595 &24.448 &0.805\\ 
							
							H &8192  &$\times$ &$\times$ &87.3M &14.135 &0.1214 &26.038 &0.908 &87.3M &23.921 &\underline{\bf 0.1553} &\underline{\bf 24.535} &0.808\\ 
							
							\qiankun{I} &\qiankun{8192} &\qiankun{$\checkmark$} &\qiankun{$\times$} &\qiankun{87.3M} &\qiankun{13.887} &\qiankun{0.1196} &\qiankun{26.196} &\qiankun{0.912} &\qiankun{87.3M} &\qiankun{24.179} &\qiankun{0.1599} &\qiankun{24.413} &\qiankun{0.808} \\ 
							
							J &8192  &$\checkmark$ &$\checkmark$ &87.3M &\underline{\bf 13.658} &\underline{\bf 0.1174} &\underline{\bf 26.321} &\underline{\bf 0.914} &87.3M &\underline{\bf 21.538} &\underline{\bf 0.1553} &24.521 &\underline{\bf 0.813}\\
							\hline
						\end{tabular}
					\label{tab: ablation_study_on_improved_pvqvae_and_transformer}
				\end{table*}

		\begin{figure}[t]
			\centering
			\includegraphics[width=1.0\columnwidth]{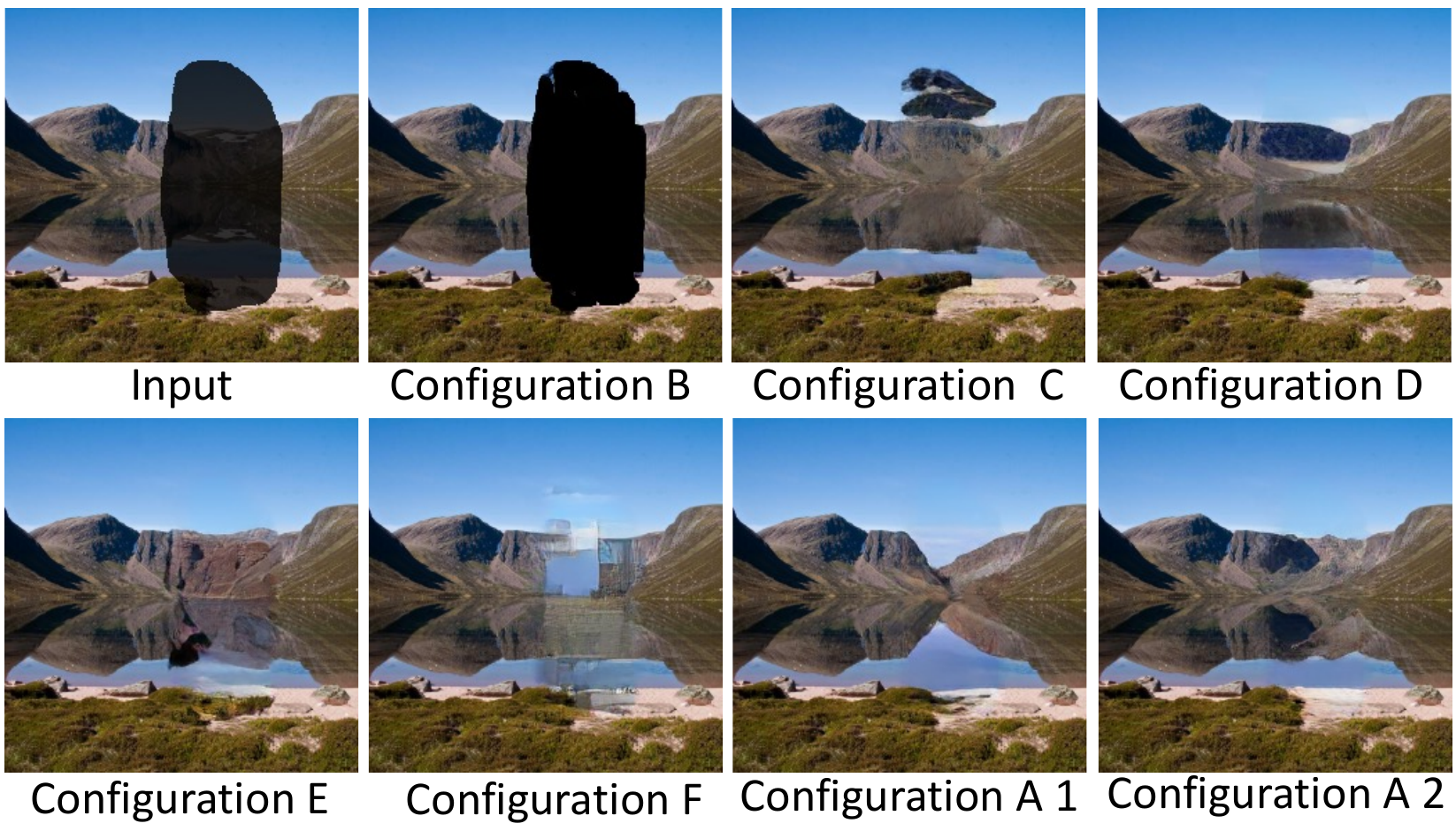} 
			\caption{Inpainted results under different configurations on Places2~\cite{zhou2017places}.} 
			\label{figure: ablation_inpainting_results}
		\end{figure}

		\subsection{Discussions}
		\label{sec: ablation_study} 
		\qiankun{In this section, we provide some discussions on the design of PUT and the potential of applying the pretrained transformer in PUT to downstream tasks. Without loss of generality, the PUT trained without conditions is evaluated.}
			
		\subsubsection{Effectiveness of Main Components} 
		\label{sec:effectiveness_of_different_components}
	We first show the effectiveness of the patch-based encoder, dual-codebook, multi-scale guided decoder and un-quantized transformer. To this end, different configurations are designed. Results are shown in \Tref{tab: ablation_study_on_put_in_conference} and \Fref{figure: ablation_inpainting_results}.

	Configuration A is the default setting in our conference paper~\cite{liu2022reduce}. We first replace the patch-based encoder with a normal CNN based encoder, which is implemented with convolution layers (Configuration B). The model performs the worst in all metrics, demonstrating the effectiveness of the non-overlapping patch partition design. Within CNN based encoder, the input images are processed in a sliding window manner, introducing the interaction between masked and unmasked regions, which is fatal to the transformer for the prediction of masked regions.

		Configuration C removes the codebook $\mathbf{e'}$ from P-VQVAE. The only difference between configurations A and C is the training of P-VQVAE with one or two codebooks since the codebook $\mathbf{e'}$ is not used in the inference stage. P-VQVAE learns more discriminative features for masked and unmasked patches with the help of dual-codebook. Interestingly, the model with codebook $\mathbf{e'}$ indeed performs better than the model without it, except for the FID and LPIPS scores on FFHQ. We speculate that face generation is much easier because all faces share a similar structure. As we can see in \Fref{figure: ablation_inpainting_results}, without the help of codebook $\mathbf{e'}$, the model sometimes predicts \emph{black} patches, which are similar to those patches containing missing pixels. By contrast, the dual-codebook helps PUT achieve overall better performance.

	\begin{figure}[t]
			\centering
			\includegraphics[width=1.0\columnwidth]{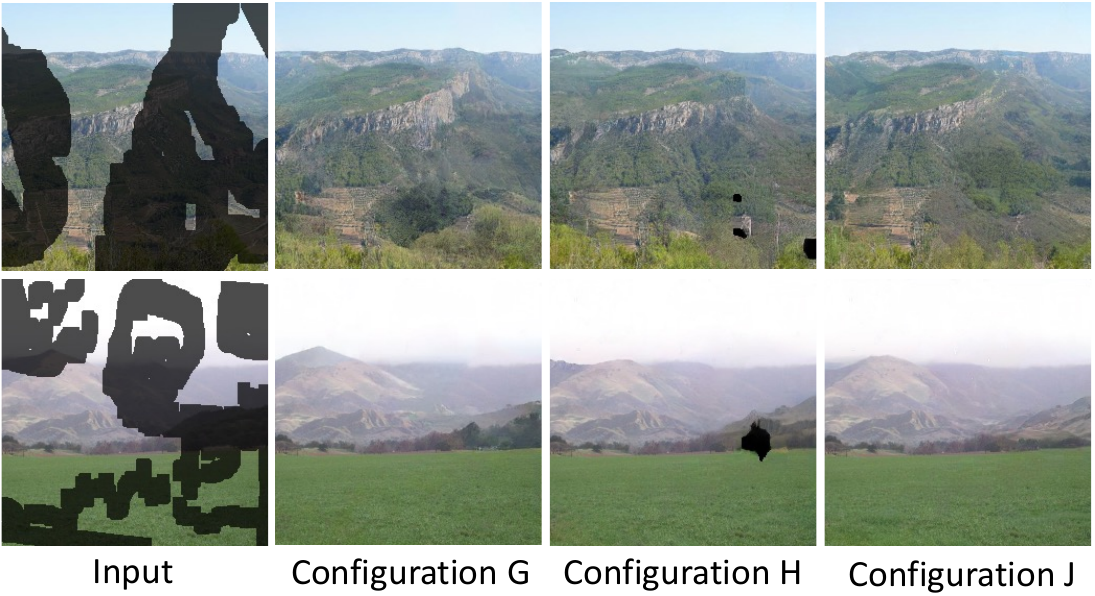} 
			\caption{Inpainted results under different configurations on Places2~\cite{zhou2017places}.  More details about these configurations please refer to \Tref{tab: ablation_study_on_improved_pvqvae_and_transformer}. Shown at $256 \times 256$.}
			\label{figure: ablation_inpainting_results_2}
		\end{figure}

	Configuration D removes the guidance of the reference branch, which means that the model constructs the inpainted image without referring to the input masked image, leading to inferior performance in terms of all metrics. With the help of multi-scale guidance from the reference branch, some useful textures can be recovered from the unmasked regions in the reference image. As we can see in \Fref{figure: ablation_inpainting_results}, the result constructed without referring to the input image is over-smoothed and unnatural.

	Configuration E feeds the transformer with quantized vectors rather than the original feature vectors from the encoder, which performs much poorer in all metrics than configuration A. Without quantizing feature vectors to discrete representations, no information loss happens in this step, which helps the transformer to understand complex content and maintain the inherent meaningful textures in the input image. However, the training of the transformer should be carefully designed by randomly quantizing the input feature vectors since only quantized vectors can be obtained for masked regions at the inference stage. The effectiveness of such random quantization during training is obvious while comparing configuration A with F.

		\begin{figure*}[t]
			\centering
			\includegraphics[width=2.07\columnwidth]{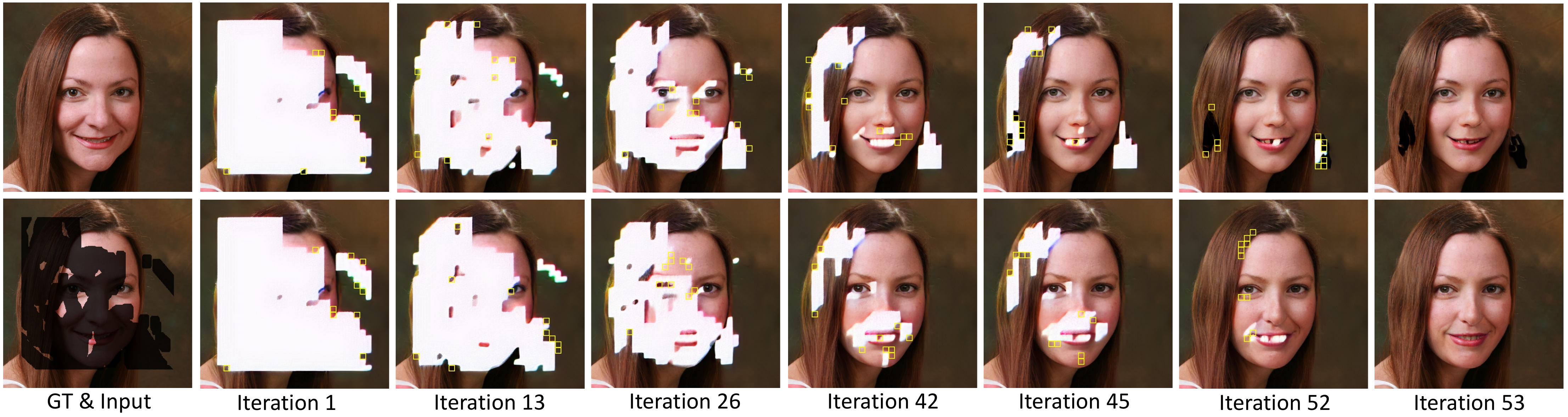}
			\caption{Top: Artifacts produced by PUT  without the help of mask embedding. Bottom: Vivid results produced by PUT with the help of mask embedding. The intermediate images are reconstructed by replacing the remanent masked tokens with a \emph{white} token for better comparison. We set $\mathcal{K}_1=10$ and $\mathcal{K}_2=200$. The patches marked by yellow boxes are those inpainted patches at that iteration. Shown at $512 \times 512$ resolution.}
			\label{figure: artifacts_and_mask_embedding}
		\end{figure*}

			\begin{table*}[t]
			\setlength{\tabcolsep}{2.3pt}
			\centering
			\caption{The impact of $\mathcal{K}_1$. Except that the last row (marked by $^\sharp$) is evaluated with $\mathcal{K}_2$=1 , others are evaluated with  $\mathcal{K}_2 = 200$.  Time consumption is averaged by the number of all images in the test split. Reported on  $256 \times 256$ resolution.} 
			\begin{tabular}{l|c|c|c|c|c|c|c|c|c|c|c|c|c|c|c}
				\hline
				Mask Ratio & \multicolumn{5}{c|}{20\%-40\%} &\multicolumn{5}{c|}{40\%-60\%} &\multicolumn{5}{c}{10\%-60\%} \\
				\hline
				\multirow{2}*{\diagbox{$\mathcal{K}_1$}{Metrics}} & \multicolumn{2}{c|}{FFHQ~\cite{karras2019style}} & \multicolumn{2}{c|}{Places2~\cite{zhou2017places}} &\multirow{2}*{\makecell{Time \\ s/image}} &\multicolumn{2}{c|}{FFHQ~\cite{karras2019style}} & \multicolumn{2}{c|}{Places2~\cite{zhou2017places}} &\multirow{2}*{\makecell{Time \\ s/image}} &\multicolumn{2}{c|}{FFHQ~\cite{karras2019style}} & \multicolumn{2}{c|}{Places2~\cite{zhou2017places}} &\multirow{2}*{\makecell{Time \\ s/image}} \\ 
				\cline{2-5}\cline{7-10}\cline{12-15}
				& FID$\downarrow$ & LPIPS$\downarrow$ & FID$\downarrow$ & LPIPS$\downarrow$ & & FID$\downarrow$ & LPIPS$\downarrow$ & FID$\downarrow$ & LPIPS$\downarrow$  &  & FID$\downarrow$ & LPIPS$\downarrow$ & FID$\downarrow$ & LPIPS$\downarrow$  \\
				\hline
				
				1    &12.186 &0.0948 &19.567 &0.1270 &7.822  &19.157 &0.1717 &30.152 &0.2222 &11.453  &13.658 &0.1174  &21.293 &0.1553 &8.849 \\
				
				5      &12.053 &0.0945 &19.654 &0.1263 &1.584 &19.691 &0.1718 &\underline{\bf 28.893} &0.2208 &2.295    &13.740 &0.1169  &21.612 &0.1544 &1.777 \\ 
				
				20    &11.891 &0.0944 & 19.028 &0.1253 &0.415 &19.458 &0.1712 &29.122 &0.2193 &0.587    &13.805 &0.1167  &21.158 &0.1532 &0.464\\ 
				
				100    &12.098 &0.0938 & 18.678 &0.1242 &0.105 & 20.402 &0.1719 & 29.130 & 0.2173 &0.137  &14.137 &0.1165  &\underline{\bf 21.075} &0.1523 &0.116 \\
				
				\hline
				
				$All$ &12.181 &0.9306 &\textbf{\underline{18.219}} &\textbf{\underline{0.1223}} &\multirow{2}*{\textbf{\underline{0.042}}} &23.510 &0.1765 &29.758 &\textbf{\underline{0.2160}} &\multirow{2}*{\textbf{\underline{0.043}}} &15.006 &0.1176 &21.234 &\textbf{\underline{0.1505}}  & \multirow{2}*{\textbf{\underline{0.043}}}\\ 
				
				$All^{\sharp}$ &\textbf{\underline{10.844}} &\textbf{\underline{0.0870}} &19.384 &0.1235 & &\textbf{\underline{18.842}} &\textbf{\underline{0.1605}} &36.566 &0.2244 & &\textbf{\underline{12.728}} &\textbf{\underline{0.1088}} &23.700 &0.1542 & \\ 
				
				\hline
			\end{tabular}
			\label{tab: impact_of_K1}
		\end{table*}

		\begin{table}[t]
			\setlength{\tabcolsep}{4pt}
			\footnotesize
			\centering
			\caption{The accuracy and probability of predicted tokens. Acc@MaxProb is the accuracy of tokens that with maximum probabilities. Prob@GT is the mean of those probabilities that correspond to  ground-truth tokens.  Evaluated at $512 \times 512$  with $\mathcal{K}_1=1$, $\mathcal{K}_2 = 200$ and mask ratio 10\%-60\%.} 
			\begin{tabular}{c|c|c|c|c}
				\hline
				& \multicolumn{2}{c}{w/o mask embedding} & \multicolumn{2}{|c}{with mask embedding} \\
				\hline 
				& Acc@MaxProb & Prob@GT & Acc@MaxProb & Prob@GT \\
				\hline
				FFHQ~\cite{karras2019style} &24.0\% &0.172 &24.4\% &0.182 \\
				\hline
				Places2~\cite{zhou2017places} &27.1\% &0.201 &28.0\% &0.220 \\
				\hline
			\end{tabular}
			\label{tab: poly_loss_and_ce_loss}
			\vspace{-5pt}
		\end{table}

		\subsubsection{Discussion on Latent Vectors and Mask Embedding}
		\label{sec:effectiveness_of_mask_embedding}
	Here, we show the effectiveness of mask embedding and the learning of more latent vectors with Gumbel-softmax relaxation. Results are shown in \Tref{tab: ablation_study_on_improved_pvqvae_and_transformer}. We first switch UQ-Transformer in configuration A to a lighter one, which is implemented based on ViT-Base~\cite{dosovitskiy2020image}. The performance drops slightly when a lighter UQ-Transformer is used.

	Interestingly, when the number of latent vectors is increased from 512 to 8192, no matter whether the Gumbel-softmax relaxation is adopted (configuration I) or not (configuration H), the model achieves better performance on FFHQ~\cite{karras2019style}, but inferior performance on Places2~\cite{zhou2017places}. However, the performance on both datasets is boosted with the help of mask embedding (configuration J). Here, we give an explanation of why the model achieves poorer performance on Places2 with more latent vectors even though the reconstruction capability of patch-based auto-encoder is greatly improved (FID/PSNR are improved from 35.183/23.560 to 33.810/24.594). The performance drop mainly comes from the artifacts: some unnatural \emph{black} pixels are easily produced when the input image contains some natural black pixels (for example, the images produced by PUT$_{\rm CVPR}$~\cite{liu2022reduce} in \Fref{figure: inpainting_results_256} on Places2). Different from photos of faces in FFHQ, textures or patterns in pictures taken from natural scenes sometimes contain scattered black pixels, for example, mountains and grasses. Such scattered black pixels are more easily recognized by PUT when more latent vectors are learned, making the transformer predict black pixels more often. Two examples are shown in  \Fref{figure: ablation_inpainting_results_2}. Such artifacts become more severe when applying PUT to higher resolutions. An example from FFHQ with resolution $512\times 512$ is shown in \Fref{figure: artifacts_and_mask_embedding}. But when mask embedding is introduced (configuration J), PUT is greatly boosted. 
		
	In \Tref{tab: poly_loss_and_ce_loss}, we show the accuracy of predicted tokens and the probability scores for ground-truth tokens. With the help of mask embedding, the transformer predicts the ground-truth tokens with a higher frequency (Acc@MaxProb) and a higher confidence (Prob@GT).

			\begin{figure}[t]
			\centering
			\subfigure[FFHQ~\cite{karras2019style}]{
				\begin{minipage}[t]{0.49\linewidth}
					\centering
					\includegraphics[width=0.99\linewidth]{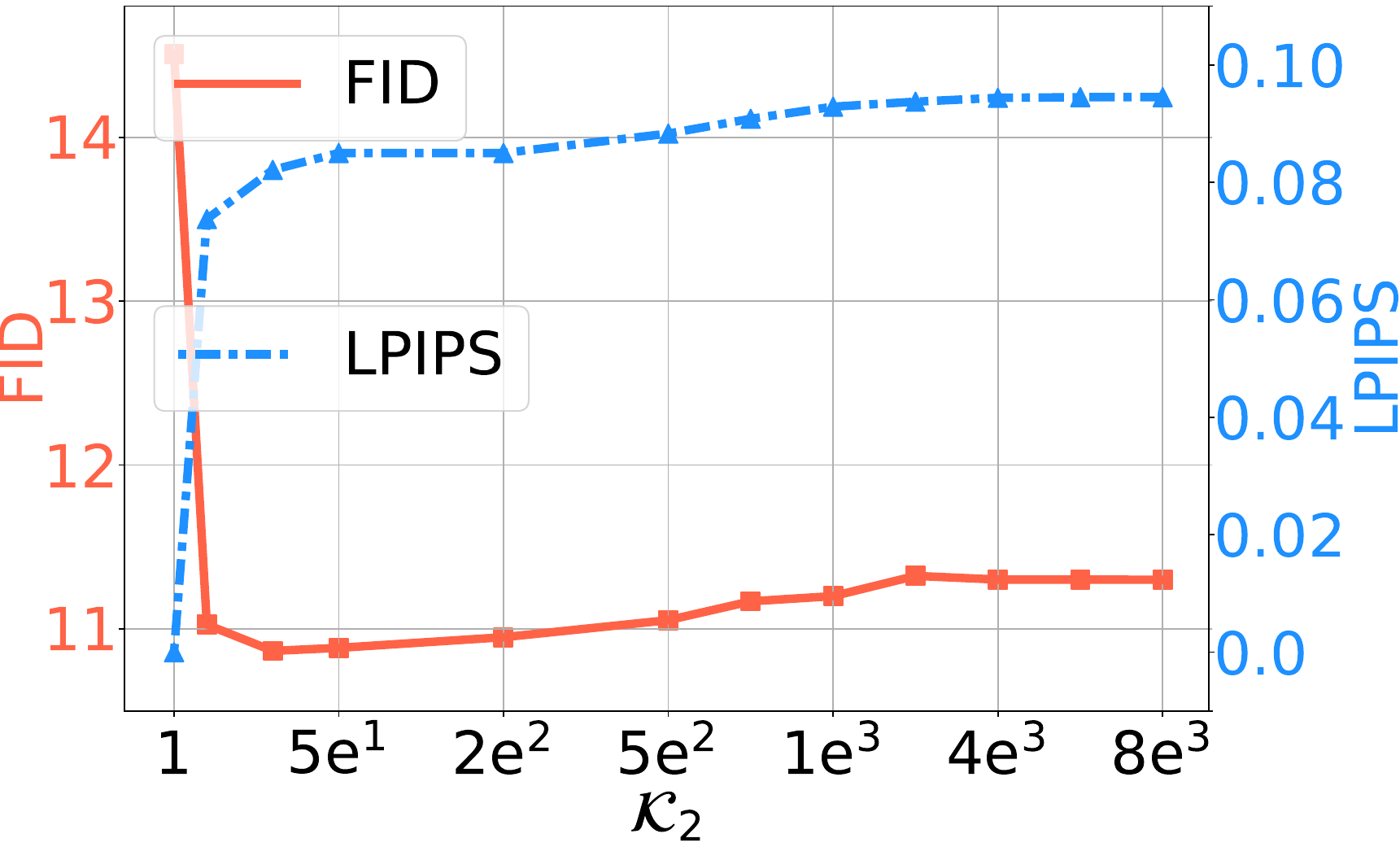}
				\end{minipage}%
			}%
			\subfigure[Places2~\cite{zhou2017places}]{
				\begin{minipage}[t]{0.49\linewidth}
					\centering
					\includegraphics[width=0.99\linewidth]{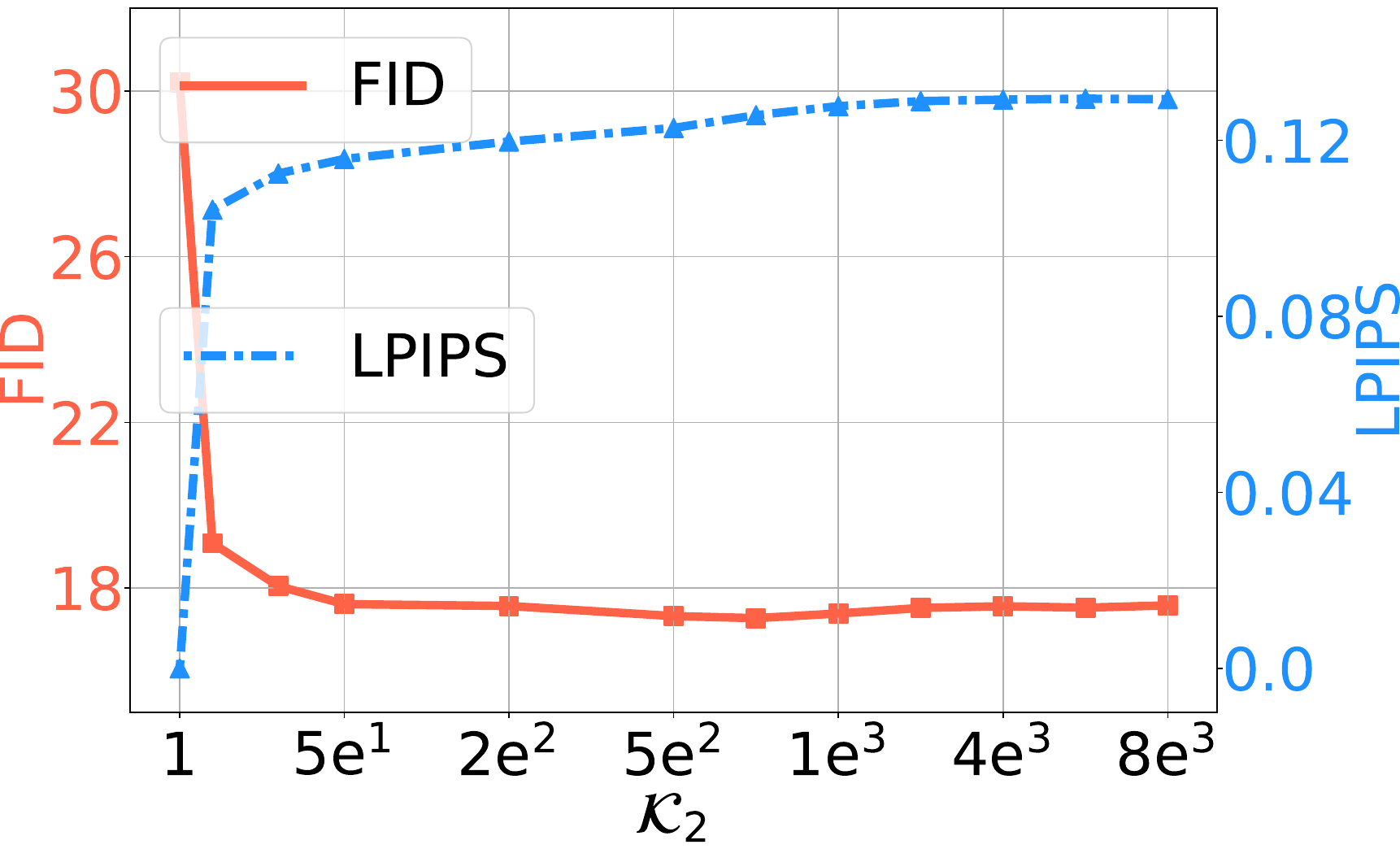}
				\end{minipage}%
			}%
			
			\centering
			\caption{LPIPS and FID curves with respect to $\mathcal{K}_2$ on different datasets when $\mathcal{K}_1 = 1$. 
				Evaluated at  $256 \times 256$ resolution with mask ratio 10\%-60\%.}
			\label{fig: lpips_and_fid_k_curve}
			\vspace{-5pt}
		\end{figure}

		\begin{table*}[t]
				\setlength{\tabcolsep}{1.5pt}
				\centering
				\mycolor
				\mycaption{Applying pretrained transformers to downstream tasks. RL: Representation learning. GM: Generative modeling.}  
				\begin{tabular}{c|cc|c|c|c|c|cc|cc}
					\hline
					\multirow{3}*{Methods} &\multicolumn{2}{c|}{Pretraining details} &Inpainting &\multicolumn{2}{c|}{Classification} &\multicolumn{5}{c}{Object detection and instance segmentation} \\
					\cline{2-11}
					&\multirow{2}*{\makecell{Task}} &\multirow{2}*{\makecell{Epochs/\\Batch size}}&ImageNet~\cite{deng2009imagenet} &\multirow{2}*{\makecell{Resolution/\\Patch size}} & ImageNet~\cite{deng2009imagenet} &\multirow{2}*{\makecell{Resolution/\\Patch size}} &\multicolumn{2}{c}{COCO~\cite{lin2014microsoft}} &\multicolumn{2}{|c}{LVIS~\cite{gupta2019lvis}} \\
					\cline{4-4} \cline{6-6} \cline{8-11}
					& & &FID$\downarrow$ & &Acc@1$\uparrow$ & &AP$^{box}$$\uparrow$ &AP$^{mask}$$\uparrow$ &AP$^{box}$$\uparrow$ &AP$^{mask}$$\uparrow$ \\
					\hline
					Random~\cite{he2022masked}  & - & - & - &224/16 &82.3 & 1024/16 &38.0 &34.7 &25.1 &24.7 \\
					\hline
					MAE (CVPR,2022)~\cite{he2022masked}  &\multirow{3}*{RL} & 1600/4096  &\underline{\bf 63.911} &224/16 &83.6 &\multirow{3}*{1024/16} &\underline{\bf 51.3} &\underline{\bf 45.7} &\underline{\bf 38.8} &\underline{\bf 36.9} \\
				    SimMIM (CVPR,2022)~\cite{xie2022simmim} & & 800/2048 &69.873 &224/16 &\underline{\bf 83.8} & &49.7 &44.2 & 36.5 &34.8 \\
					BEiT (ICLR,2022)~\cite{bao2021beit} & & 800/2048 &127.913 &224/16 &83.2 & &48.4 &43.0 &35.8 &34.3 \\
					\hline
					MAGE (CVPR,2023)~\cite{li2023mage} & RL + GM &1600/4096 &145.952 &256/16 &82.5 &\multirow{3}*{1024/16} &46.0 &40.1 &32.9 &29.7 \\
					PUT, $512\times512$, (Ours) &\multirow{2}*{Inpainting} &300/384 &25.452 &224/16 &82.7 & &48.1 &42.5 &34.5 &32.6\\
					PUT, $256\times256$, (Ours) &  & 300/384 &\underline{\bf 24.794} &224/16 & \underline{\bf 82.9} & &\underline{\bf 49.0} &\underline{\bf 43.5} &\underline{\bf35.4} &\underline{\bf33.3} \\
					
					\hline
					
				\end{tabular}
				\label{tab: downstream_tasks}
			\end{table*}

		\subsubsection{Discussions on Sampling Strategy}
		\label{sec: dicussion_on_sampling_strategy}
	The impact of $\mathcal{K}_2$ when $\mathcal{K}_1 = 1$ is shown in \Fref{fig: lpips_and_fid_k_curve}. To get the diversity of inpainted images, five pairs of results for each input are generated to calculate the mean LPIPS~\cite{zhang2018unreasonable} score. Meanwhile, the fidelity between all generated results and ground-truth images is also obtained. As we can see, when $\mathcal{K}_2$ is large enough ($\mathcal{K}_2 \ge 50$), the diversity and fidelity are saturated, which means PUT is not sensitive to the value of $\mathcal{K}_2$. However, when $\mathcal{K}_2$ is too small ($\mathcal{K}_2 < 10$), PUT loses the capability of diversity and fidelity. We set $\mathcal{K}_2=200$ as default to balance the diversity and fidelity.

	To evaluate the influence of $\mathcal{K}_1$, we generate one result for each input. The LPIPS and FID scores between generated images and ground-truth images are calculated. Results are shown in \Tref{tab: impact_of_K1}. We can find that:
	1) The time consumption is less if PUT takes fewer iterations (\textit{i.e.}, with a larger $\mathcal{K}_1$).
	2) The impact of $\mathcal{K}_1$ on the quality of inpainted images is limited, which means PUT is somewhat robust to different values of $\mathcal{K}_1$.
	3) When inpainting an image within one iteration and setting $\mathcal{K}_2$ to 200 (denoted as \emph{All}), PUT achieves worse performance on FFHQ when the mask ratio is large (\textit{e.g.}, 40\%-60\%). However, such a phenomenon is not observed on Places2. The reason is that the faces in FFHQ are more structured than natural scene pictures in Places2. When the tokens for masked patches are sampled from top-200 elements independently, the structure of faces is destroyed.
	4) In contrast to the aforementioned phenomenon, when the tokens for masked patches are sampled at one iteration with $\mathcal{K}_2 = 1$ (denoted as $All ^\sharp$), PUT performs much better on FFHQ but much worse on Places2. The reason is that the inpainted images are over-smoothed. Such smoothness is acceptable for faces but unnatural for natural scene pictures.

	For iteratively image inpainting, we set $\mathcal{K}_1=20$, $\mathcal{K}_2=200$ for different datasets. For image inpainting within one iteration ($\mathcal{K}_1=All$), we set $\mathcal{K}_2=1$ for FFHQ and ImageNet and $\mathcal{K}_2=200$ for Places2.

		\qiankun{\subsubsection{Comparison with Masked Image Modeling Methods}}
	\qiankun{Finally, we compare PUT with some popular MIM methods for representation learning, including MAE~\cite{he2022masked}, BEiT~\cite{bao2021beit}, SimMIM~\cite{xie2022simmim} and MAGE~\cite{li2023mage}. Among them, MAGE tries to unify generative modeling with representation learning. Results are shown in \Tref{tab: downstream_tasks}. For inpainting, the masked images (with a mask ratio ranging from 10\%-60\%) are fed into the pretrained models. For classification, the results of compared methods are extracted from their papers while the results of PUT are obtained through finetuning it on ImageNet for 200 epochs. For object detection and instance segmentation, all results are produced by us using the codes of ViTDet~\cite{li2022exploring} through optimizing different methods for 50 epochs. The first kernel in $256 \times 256$ pretrained PUT model is interpolated to adapt to handle $16 \times 16$ image patches.}
	
	\qiankun{Overall, the methods that have image synthesis capability, \textit{e.g.}, PUT and MAGE, are inferior to those methods that are dedicated to representation learning when applied to downstream tasks. However, PUT is much superior to MAGE and comparable to BEiT even though it is designed for image inpainting rather than representation learning. For the task of image inpainting, MAGE and BEiT perform much worse than other methods. The reason is that similar convolution-based auto-encoders in configuration B (Ref. \Tref{tab: ablation_study_on_put_in_conference}) are adopted to provide the quantized discrete tokens~\cite{he2022masked, xie2022simmim}, demonstrating the effectiveness of P-VQVAE for image inpainting.}

	\section{Conclusions and Limitations}
	\label{sec: conclusion} 
	In this paper, we present a novel method, PUT, for pluralistic image inpainting. PUT consists of two main components: 1) patch-based auto-encoder (P-VQVAE) and 2) un-quantized transformer (UQ-Transformer). With the help of P-VQVAE and UQ-Transformer, PUT processes the original high-resolution image without quantization. Such practice preserves the information contained in the input image as much as possible. Experimental results demonstrate the superiority of PUT, including fidelity and diversity, especially for large masked regions and complex scenes.

	\qiankun{The main limitations of PUT include: 1) Inference time for generating diverse results. Though the inference time has been greatly reduced with the multi-token sampling strategy, PUT still needs several iterations to inpaint one image. However, it is a common issue of existing transformer based autoregressive methods~\cite{wan2021high, ramesh2021zero, esser2021taming}. 2) Generalization of image resolution. UQ-Transformer is sensitive to the length of sequence since it is trained with a fixed number of image patches and position embeddings. It may be solved by dynamically changing the resolution of input images during the training stage and using more flexible position embeddings~\cite{vaswani2017attention}. We leave these to our future work.}

	\vspace{10pt}
\noindent\textbf{Acknowledgements:}
\revise{This work was supported by the National Natural Science Foundation of China (62331006, 62171038, and 62088101), and the Fundamental Research Funds for the Central Universities.}
	
		\ifCLASSOPTIONcaptionsoff
		\newpage
		\fi

		\bibliographystyle{IEEEtran}
		\bibliography{cite.bib}

	\end{document}